\documentclass[10pt,twocolumn,letterpaper]{article}

\usepackage{cvpr}
\usepackage{tikz}
\usepackage{array}
\usepackage{colortbl}
\usepackage{boldline}
\usepackage{comment}
\usepackage{multirow}
\usepackage{tabularx}
\usepackage{algorithm2e}
\usepackage{algpseudocode}
\usepackage{mathtools}
\usepackage{amssymb}
\usepackage{arydshln}
\usepackage{dblfloatfix}
\usepackage[accsupp]{axessibility}

\definecolor{cvprblue}{rgb}{0.21,0.49,0.74}

\definecolor{rblue}{HTML}{2676b8}
\definecolor{blue}{HTML}{2676b8}
\definecolor{rred}{HTML}{d11e1f}
\definecolor{red}{HTML}{d11e1f}
\definecolor{rgreen}{HTML}{289049}
\definecolor{rorange}{HTML}{dd4d04}

\usepackage[pagebackref,breaklinks,colorlinks,citecolor=cvprblue]{hyperref}

\usepackage{glossaries}
\glsdisablehyper

\newacronym{ai}{AI}{Artificial Intelligence}
\newacronym{dl}{DL}{Deep Learning}
\newacronym{dnn}{DNN}{Deep Neural Network}
\newacronym{lrp}{LRP}{Layer-wise Relevance Propagation}
\newacronym{lsb}{LSB}{least significant bit}
\newacronym{xai}{XAI}{eXplainable Artificial Intelligence}
\newacronym{crp}{CRP}{Concept Relevance Propagation}
\newacronym{amax}{ActMax}{Activation Maximization}
\newacronym{rmax}{RelMax}{Relevance Maximization}
\newacronym{auc}{AUC}{Area Under Curve}
\newacronym{aoc}{AOC}{Area Over Curve}
\newacronym{conv}{Conv}{convolutional}
\newacronym{svm}{SVM}{Support Vector Machine}
\newacronym{roi}{ROI}{Region of Interest}
\newacronym{lcrp}{L-CRP}{CRP for Localization Models}
\newacronym{rrr}{RRR}{Right for the Right Reason}
\newacronym{cdep}{CDEP}{Contextual Decomposition Explanation Penalization}
\newacronym{clarc}{ClArC}{Class Artifact Compensation}
\newacronym{aclarc}{\mbox{A-ClArC}}{Augmentive ClArC}
\newacronym{pclarc}{\mbox{P-ClArC}}{Projective ClArC}
\newacronym{rclarc}{\mbox{R-ClArC}}{Reactive ClArC}
\newacronym{rrclarc}{RR-ClArC}{Right Reason ClArC}
\newacronym{ml}{ML}{Machine Learning}
\newacronym{cse}{CSE}{complete skin examination}
\newacronym{cav}{CAV}{Concept Activation Vector}
\newacronym{tcav}{TCAV}{Testing with CAV}
\newacronym{spray}{SpRAy}{Spectral Relevance Analysis}
\newacronym{iterrev}{IterRev}{Iteratively Revealing and Revising Spurious Model Behavior}
\newacronym{r2r}{R2R}{Reveal to Revise}
\newacronym{xil}{XIL}{eXplanatory Interactive Learning}
\newacronym{sem}{SEM}{Standard Error of the Mean}
\newacronym{se}{SE}{Standard Error}

\newtheorem{definition}{Definition}[section]

\makeatletter
\DeclareRobustCommand\onedot{\futurelet\@let@token\@onedot}
\def\@onedot{\ifx\@let@token.\else.\null\fi\xspace}

\def\eg{\emph{e.g}\onedot} 
\def\ie{\emph{i.e}\onedot} 
 
\def\etc{\emph{etc}\onedot}

\newcolumntype{R}{>{\raggedleft\arraybackslash}X}
\newcolumntype{L}{>{\raggedright\arraybackslash}X}
\newcolumntype{Y}{>{\centering\arraybackslash}X}
\newcolumntype{A}{ >{$} R <{$} @{} >{${}} L <{$} }
\newcommand{\specialcell}[2][c]{%
  \begin{tabular}[#1]{@{}c@{}}#2\end{tabular}}

\title{Reactive Model Correction: Mitigating Harm to Task-Relevant Features via Conditional Bias Suppression}

\author{
Dilyara Bareeva$^{1}$,
Maximilian Dreyer$^{1}$,
Frederik Pahde$^{1}$,
Wojciech Samek$^{1,2,3,\dagger}$,
Sebastian Lapuschkin$^{1,\dagger}$\\
$^1$ Fraunhofer Heinrich Hertz Institute,
$^2$ Technical University of Berlin, \\
$^3$ BIFOLD – Berlin Institute for the Foundations of Learning and Data\\
$^\dagger${\small corresponding authors:}
{\tt\small  \{wojciech.samek\,|\,sebastian.lapuschkin\}@hhi.fraunhofer.de}
}

\begin{document}
\maketitle
\begin{abstract}
Deep Neural Networks are prone to learning and relying on spurious correlations in the training data, which, for high-risk applications, can have fatal consequences. Various approaches to suppress model reliance on harmful features have been proposed that can be applied post-hoc without additional training. Whereas those methods can be applied with efficiency, they also tend to harm model performance by globally shifting the distribution of latent features. To mitigate unintended overcorrection of model behavior, we propose a reactive approach conditioned on model-derived knowledge and eXplainable Artificial Intelligence (XAI) insights. While the reactive approach can be applied to many post-hoc methods, we demonstrate the incorporation of reactivity in particular for P-ClArC (Projective Class Artifact Compensation), introducing a new method called R-ClArC (Reactive Class Artifact Compensation). Through rigorous experiments in controlled settings (FunnyBirds) and with a real-world dataset (ISIC2019), we show that introducing reactivity can minimize the detrimental effect of the applied correction while simultaneously ensuring low reliance on spurious features. 
\end{abstract}    
\section{Introduction}
\label{sec:intro}

\begin{figure}[t]
  \includegraphics[width=\columnwidth]{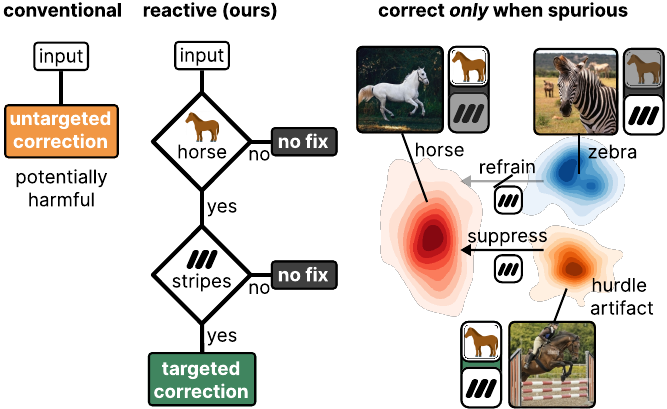}
  \centering
  \caption{Reactive Model Correction: Whereas traditional post-hoc model correction approaches are applied to all samples \emph{uniformly},
  we propose \emph{conditional} suppression of artifacts. One possible condition for triggering correction is the combination of a specific class prediction and the presence of a spurious feature (\emph{left}).
  This prevents the suppression of concepts when unnecessary or even harmful: 
  When correcting, \eg, for a ``hurdle''-artifact (related to ``stripe'' features), 
  we refrain from suppression for zebra samples,
  as stripe textures are now valid discriminative features and crucial to discerning zebras from horses (\emph{right}).}

\label{fig:title_figure}
\end{figure}

Modern \gls{dnn} architectures yield remarkable results for a plethora of complex tasks, including high-stake applications, such as medicine~\cite{briganti2020artificial}, finance~\cite{rouf2021stock}, or criminal justice~\cite{travaini2022machine}.
However, it has been shown that \glspl{dnn} are at risk of learning shortcuts based on spurious correlations due to imperfections in the available training data~\cite{lapuschkin_unmasking_2019,geirhos_shortcut_2020,anders_finding_2022,neuhaus2023spurious}, compromising the reliability of these models in high-risk scenarios. 
Some notable examples include melanoma detection models using visible band-aids as evidence \emph{against} cancerous melanoma, as they only occurred next to benign lesions in the training data~\cite{rieger2020interpretations}, or bone age prediction models exploiting the spurious correlation between the size of lead markers and bone age caused by the specifics of data processing~\cite{pahde_reveal_2023}. Another example is the usage of hurdles as features for horse classification~\cite{lapuschkin2019unmasking}, as illustrated, \eg, in \cref{fig:title_figure} (\emph{bottom right}).

To address these model weaknesses arising from compromised training data,
a multitude of model correction approaches have been suggested in recent years to unlearn undesired model behavior, either by model re-training on modified data~\cite{wu2023discover,weng2023fast},  model fine-tuning~\cite{ross2017right,rieger2020interpretations,schramowski2020making,dreyer2023hope}, or by post-hoc model editing~\cite{de2021editing,ravfogel-etal-2020-null,santurkar_editing_2021,anders_finding_2022,LINHARDT2024pruning}.
Whereas the former two groups of approaches require access to the training data and are expensive in terms of computational resources, therefore often being infeasible to apply for large models, the latter group provides a cheap solution to modify existing models before deployment, or even during inference time. 
Commonly,
post-hoc model correction approaches model the shortcut concept as a linear direction or subspace in latent space, and modify latent representations or model parameters such that predictions become invariant towards the modeled artifact direction.

However, while such approaches successfully reduce the impact of data artifacts on model predictions, they tend to negatively impact the model performance for ``clean'' data samples without artifacts, as spurious concepts are suppressed globally (\ie, \emph{always}) due to a constant modification of the model parameters or the elimination of latent subspaces.
When artifact representations are modeled inaccurately or entangled with other features,
artifact suppression can lead to unintended and harmful suppression of important task-relevant features~\cite{NEURIPS2022_725f5e80}. 
For instance, when suppressing the direction encoding the ``hurdle'' concept in horse detectors, it also affects related concepts like the direction for ``stripes'', potentially impacting the model's ability to differentiate between horses and zebras.

\begin{table*}[ht]
  \centering
    \caption{Examples of reactive conditions suitable for reactive model correction and their respective informing XAI methods.}
  \renewcommand{\arraystretch}{1.3} 
    \begin{tabular}{@{}p{0.03\textwidth}p{0.16\textwidth}@{\hspace{2.5em}}p{0.42\textwidth}@{\hspace{2.5em}}p{0.21\textwidth}p{0.01\textwidth}}
  \toprule
    & Reactive Condition & Description & Informing XAI Methods &\\
    \midrule
    \rotatebox[origin=c]{360}{\begin{tikzpicture}[font=\small,overlay]
      \draw[->, line width=0.5pt] (0.5,0.35) -- (0.5,-6.5);
      \node[anchor=south, color=rred, rotate=90] at (0.5,-0.6) {less granular};
      \node[anchor=south, color=rgreen, rotate=90] at (0.5,-5.4) {more granular};
    \end{tikzpicture}}
    & Predicted class &Correction is only applied when a class that is associated with an artifact is predicted. & none needed &     \rotatebox[origin=c]{360}{\begin{tikzpicture}[font=\small,overlay]
      \draw[<->, line width=0.5pt] (-0.05,0.35) -- (-0.05,-1.5);
      \node[anchor=south, color=rblue, rotate=270] at (-0.05,-0.6) {covered};
    \end{tikzpicture}}
    \\

    & Artifact presence & Correction is only initiated if the presence of the artifact has been detected in the sample. & \gls{cav}~\cite{kim_interpretability_2018} & \\    
    \cmidrule{2-4}

    & Artifact relevance &  Correction is only initiated if the artifact has been identified as relevant for the prediction. & attributions for input~\cite{10.1371/journal.pone.0130140} or latent~\cite{achtibat_attribution_2023,fel_holistic_2023} features & \\

    & Prediction strategy & A sample is categorized into one of the known prediction sub-strategies (clusters), \eg based on activations and attributions of hidden units. The reactive condition is triggered, if the cluster has been previously identified as undesirable. & PCX\cite{dreyer_understanding_2023}, strategic clusters\cite{fel_holistic_2023} & \\
    & Domain-expert-in-the-loop & Triggered manually by a domain expert upon inspection of the information provided by XAI methods, including heatmap visualizations, active and identified prediction strategies, \etc & All of the above & \\
    \bottomrule
  \end{tabular}
  \label{react_cond}
\end{table*}

In order to minimize ``collateral damage'', we introduce a \emph{reactive} model correction paradigm, where corrective measures are applied during the inference only if certain conditions are met.
Such conditions can include the prediction of a certain class and/or presence and relevance of the concept to the model decision-making.
To that end, we leverage insights from local and global \gls{xai} methods to understand and recognize the role of certain concepts (\eg, data artifacts) for a single prediction and reactively update the prediction if required, while leaving other predictions unaffected, as illustrated in Fig.~\ref{fig:title_figure}.

We empirically demonstrate the benefits of our approach compared to non-reactive post-hoc model corrections using the ISIC2019~\cite{8363547,combalia2019bcn20000, mitsuhara2019embedding} and FunnyBirds~\cite{hesse_funnybirds_2023} datasets, with both controlled and real-world data artifacts.
Our contributions include the following:
\begin{enumerate}
    \item We introduce \emph{reactive} model correction approach that \emph{only} corrects model behavior when required, \eg, for a specific class and/or when an artifact is predicted to be present.
    \item We demonstrate the entanglement of concept representations of artifacts and non-artifacts.
    \item We show the superior performance of reactive model correction in quantitative evaluation using controlled and real-world data artifacts.
\end{enumerate}

\section{Related Work}
\label{sec:formatting}

Methods for model correction either require full retraining,
fine-tuning or can be applied post-hoc.
Whereas full retraining is usually necessary for methods that change the data distribution~\cite{teso2019explanatory,wu2023discover,weng2023fast},
fine-tuning often involves the regularization of a model by applying dedicated loss terms~\cite{ross2017right,rieger2020interpretations,dreyer2023hope}.
However, the creation of a representative training data set often comes with its own set of challenges, as the data cleaning process is not always straightforward and involves significant manual labor~\cite{liang2022advances}, and can introduce new biases~\cite{dodge2021documenting}. 
(Partial) retraining may be infeasible due to high computational costs and can exacerbate unlabeled shortcuts~\cite{li2023whac}.
In light of the challenges, post-hoc correction methods~\cite{NIPS2016_a486cd07,santurkar_editing_2021,belrose2023leace,ravfogel-etal-2020-null}, including the \gls{pclarc}~\cite{anders_finding_2022} method, represent a preferable alternative. Theoretical and empirical analyses presented in~\cite{NEURIPS2022_725f5e80} demonstrate that post-hoc methods, specifically INLP~\cite{ravfogel-etal-2020-null}, have the potential to eliminate not only spurious but also task-relevant features. In this work, we introduce a general framework to address this issue by applying post-hoc model correction only under specific conditions, minimizing collateral damage caused by the suppression of unintended, correlated concepts.

\section{Reactive Model Correction Framework}

Let $f: \mathcal{X} \rightarrow \mathbb{R}^D$ be a neural network, mapping input samples $x \in \mathcal{X}$ to an output for $D$ labels. 
Given a hidden layer $l$ of network $f$ with $m$ neurons, we define the \emph{feature extractor} $a:  \mathcal{X} \rightarrow \mathcal{A} \subseteq \mathbb{R}^m$ providing the latent activations of layer $l$, and the \emph{model head} $\Tilde{f}: \mathcal{A} \rightarrow \mathbb{R}^D$, mapping activations to the output.

We further introduce the notion of \emph{artifacts} as a set $\mathcal{C}$ of concepts with a size $|\mathcal{C}| = n$ that are present in the data but should not be relied upon by the network $f$ (\eg, watermarks or hurdles). For each artifact $c_i \in \mathcal{C}$, $i \in \{1, ...., n\}$, we are provided with sets of samples with the concepts present ${X^{+}_1, X^{+}_2, \dots, X^{+}_n}$ (\emph{positive} examples) and corresponding sets without the concepts ${X^{-}_1, X^{-}_2, \dots, X^{-}_n}$ (\emph{negative} examples). For each set, corresponding activation vectors $\mathcal{A}_i^{+} = \{a(x) \in \mathbb{R}^m| x \in X_i^{+}\} \subseteq \mathcal{A}$ and $\mathcal{A}_i^{-} = \{a(x) \in \mathbb{R}^m| x \in X_i^{-}\} \subseteq \mathcal{A}$ are obtained. A wide-spread assumption is that concepts learned and used by a network are encoded as Concept Activation Vectors (\gls{cav}s), \ie characteristic directions or subspaces in the latent space $\mathcal{A}$~\cite{kim_interpretability_2018}. Let $v_i$ denote a \gls{cav} for concept $c_i \in C$ pointing from $\mathcal{A}_i^{-}$ to $\mathcal{A}_i^{+}$.

\subsection{Reactive Model Correction} \label{reactive_framework}

The overall objective of model correction is to \emph{suppress} the influence of artifacts on the model decision-making. Post-hoc model correction methods commonly aim to \emph{globally} and \emph{simultaneously} address the set of all known artifacts, leading to a two-fold problem:
Firstly,
a concept might be considered spurious for one class, but encodes a valid strategy for another, as, \eg, the ``stripes'' concept in \cref{fig:title_figure} (\emph{right}) represents a valid feature for zebras but a shortcut artifact for horses.
Secondly,
the artifact concept might be entangled with other useful features in latent space.
Consequently,
artifact suppression also leads to a distorted representation of valid features.
Both problems potentially harm model performance (as measured in \cref{concepts_related}), 
and motivate a \emph{conditional}, more targeted application of bias suppression.

To address these challenges,
we propose \emph{reactive} (post-hoc) model correction. 
Essentially, this paradigm aims to initiate model correction only under specific conditions. 
In \cref{react_cond} we present an overview of potential reactive conditions,
including, \eg, the prediction of a specific class, the actual presence of an artifact, or a decision by an expert,
as well as \gls{xai} methods that can be used to measure whether the pre-defined conditions are met.

Let us formalize a condition (as, \eg, given in \cref{react_cond}) as a condition-generating function:

\begin{definition}[Condition-generating function]
Given a network $f$ and the set $\mathcal{C}$ of known artifact concepts, we define a condition-generating function $r \colon \mathcal{X} \to \mathcal{C}^{*}\subseteq \mathcal{C}$ as a function, that, for any given input, produces a (reduced) set $\mathcal{C}^{*}$ of artifacts to be removed.
\end{definition}

The core principle of the reactive model correction framework involves the identification of the artifacts to be corrected for a given sample, based on the condition-generating function. This typically involves a forward pass followed by the computation of the condition-generating function, potentially complemented by expert evaluation. Once the artifacts are determined, if any, a subsequent (partial) forward pass is performed with model correction applied specifically to address these identified artifacts.

While many post-hoc methods, including classifier editing~\cite{santurkar_editing_2021}, LEACE~\cite{belrose2023leace}, INLP~\cite{ravfogel-etal-2020-null}, and \gls{pclarc}~\cite{anders_finding_2022}, can be employed for model correction in the final step, in this work we demonstrate and evaluate our reactive approach within the \gls{pclarc} framework.

\subsection{Reactive ClArC} \label{r-clarc}

In the subsequent section, we introduce a reactive variant of \gls{pclarc}, an inference-time model correction method suppressing artifact signals modeled as linear direction in latent space.
Two core assumptions constitute the foundation of the \gls{pclarc} method. Firstly, it is assumed that introducing an artifact to a sample in the input domain leads to an increase in the activations along the corresponding \gls{cav} direction in the latent space~\cite{kim_interpretability_2018}.

Originally,
linear \emph{classifiers}, such as \glspl{svm}~\cite{cortes_support-vector_1995}, are trained on latent features to estimate \glspl{cav}.

Recently,
Pahde \etal~\cite{pahde_navigating_2024} have demonstrated that pattern-based \glspl{cav} yield more precise concept representations and therefore, when incorporated into the ClArC framework, superior performance compared to linear classifiers. The second assumption posits that all other concepts are encoded in directions orthogonal to the artifact direction. Consequently, this further implies that there is no variance in the artifact direction for all non-artifact samples.

These two assumptions lead to the \mbox{\gls{pclarc}}
\emph{backward} artifact model which suppresses the artifact features in the latent activations.

\begin{definition}[\gls{pclarc}]
Let $c_i \in \mathcal{C}$ be an artifact, $v_i$ its respective \gls{cav}, and $z_i^{-} = \frac{1}{|X_i^{-}|} \sum_{z \in X_i^{-}} a(z)$ the mean activation over non-artifact samples. 
Then, the \gls{pclarc} artifact backward model $h: \mathcal{A} \rightarrow \mathcal{A}$ for an activation vector $a_x = a(x)$ is defined as follows:
\begin{align}  \label{eq:p-clarc}
h (a_x, c_i) &= a_x - v_i v_i^T (a_x - z_i^{-}).
\end{align} 
\end{definition}

The backward artifact model $h$ effectively corrects the output of the feature extractor $a$ during the inference process, before the model head $\Tilde{f}$ is applied. We can extend this formulation to accommodate the correction of multiple artifacts simultaneously by defining an \emph{artifact subspace} as the space spanned by all artifact \gls{cav}s. We then project onto the subspace orthogonal to this artifact subspace:

\begin{definition}[Multi-Artifact \gls{pclarc}]
Let $\mathcal{C}' \subseteq \mathcal{C}$ represent a subset of artifacts with a size $|\mathcal{C}'| = k$, $1 \leq k \leq n$. Let $V_{\mathcal{C}'} =  \left[ v_i \right]_{c_i \in \mathcal{C}'}$ be the matrix comprised of the respective \gls{cav}s as column vectors. Let $Z_{\mathcal{C}'}^{-} = \bigcap_{{c_i \in \mathcal{C}'}} X^{-}_{i}$ be the intersection of negative examples, and $z_{\mathcal{C}'}^{-} = \frac{1}{|Z_{\mathcal{C}'}^{-}|} \sum_{z \in Z_{\mathcal{C}'}^{-}} a(z)$. Then, the multi-artifact \gls{pclarc} artifact backward model $\Tilde{h}: \mathcal{A}  \times\mathcal{C} \rightarrow \mathcal{A}$  for an activation vector $a_x = a(x)$ is defined as follows:

\begin{align} \label{eq:mp-clarc}
\Tilde{h} (a_x, \mathcal{C}') &= a_x - V_{\mathcal{C}'} (V_{\mathcal{C}'}^T V_{\mathcal{C}'})^{-1}V_{\mathcal{C}'}^T (a_x - z_{\mathcal{C}'}^{-}).
\end{align} 
\end{definition}

with the derivation given in \cref{mult-cav-pclarc-derivation}.

While the \gls{pclarc} method performs effectively under the given assumptions, these conditions are rarely encountered in practice, as motivated in Fig.~\ref{fig:title_figure}. 
This is confirmed by experiments in \cref{concepts_related}, illustrating that artifact \gls{cav}s exhibit strong correlations with other concept directions. 
Consequently, suppressing an artifact can inadvertently alter the representations of other potentially valid and important concepts, whether they are present or absent from a sample, thus impacting decision-making processes. Therefore, we suggest integrating reactivity into the \gls{pclarc} framework. In particular, we apply a condition-generating function to identify the set of artifacts to be suppressed prior to applying the \gls{pclarc} correction: 

\begin{definition}[R-ClArC] 
For a condition-generating function $r$ and  an activation vector $a_x = a(x)$, we define the \gls{rclarc} model $h_{\text{r}}: \mathcal{A} \rightarrow \mathcal{A}$ as follows:

\begin{align} \label{eq:r-clarc}
h_{\text{r}} (a_x) &=   \begin{cases}
      \Tilde{h}(a_x, r(x)), & r(x) \neq \varnothing \\
      a_x, & \text{otherwise}
    \end{cases}.
\end{align} 
\end{definition}

In the following, we discuss and evaluate the detailed implementation of \gls{rclarc} for two reactive conditions outlined in \cref{react_cond}: 
\emph{class-conditional} \gls{rclarc} based on \emph{label prediction} condition-generating functions,
and \emph{artifact-conditional} \gls{rclarc} given by \emph{artifact presence} condition-generating functions.

\begin{definition}[Class-condition-generating function]
Assume that for every artifact $c \in \mathcal{C}$, we are provided with a set of output labels $\mathcal{R}_c \subseteq [D] = \{1, 2, \dots, D\}$. A class-condition-generating function is then as follows:

\begin{align} \label{eq:class-cond}
r(x) = \{ c \in \mathcal{C} \mid \arg \max_{d \in [D]} f^{(d)}(x) \in \mathcal{R}_c\},
\end{align} 
\end{definition}
where $\arg \max_{d \in [D]} f^{(d)}(x)$ corresponds to the predicted output class of sample $x$.

\begin{definition}[Artifact-condition-generating function]
Assume that for every artifact $c \in \mathcal{C}$, we are provided with a binary classifier $t_c: \mathcal{X} \rightarrow \{0, 1\}$ that outputs 1 if the artifact $c$ is present in a sample $x\in\mathcal{X}$, and 0 otherwise. An artifact-condition-generating function is then as follows:

\begin{align}  \label{eq:art-cond}
r(x) = \{ c \in \mathcal{C} \mid t_c (x) = 1\}.
\end{align} 
\end{definition}

In \cref{pseudo}, we present pseudocode outlining the algorithms for \gls{pclarc} and \gls{rclarc}, and in \cref{3d} we provide a 3D toy example to illustrate these methods.

\section{Experiments}
    We address the following research questions:
    \begin{enumerate}
        \item \textbf{(Q1)} What is the degree of \emph{dependence} between representations of artifact and non-artifact concepts within models? (\cref{concepts_related})
        \item \textbf{(Q2)} How do the \emph{effectiveness} and the degree of \emph{collateral damage} caused by the reactive approach compare to the traditional model correction approach? (\cref{performance})
        
    \end{enumerate}
    
\subsection{Experimental Details}
We conduct experiments in two controlled settings using toy datasets and in one setting utilizing real-world artifacts in a benchmark dataset. 
In the first controlled setup, we generate a synthetic FunnyBirds dataset~\cite{hesse_funnybirds_2023}, consisting of two bird classes. 
We insert a \emph{backdoor} \cite{gu2019badnets} artifact (``green box'', see \cref{fig:generated}) into 33\% of the training samples originating from class 0 and flip their labels to class 1. This training configuration forces the model always to predict class 1 in the presence of the inserted ``green box'' object.

In the second setup, we create a FunnyBirds dataset comprising ten classes. We randomly select ten different background object artifacts and insert a random number of these artifacts into 50\% of training samples belonging to class 0. Since the artifacts are exclusively present in class 0, we expect a model trained on this dataset to utilize them as \emph{shortcuts} \cite{geirhos_shortcut_2020} for class 0.

For the real-world dataset, we utilize ISIC2019~\cite{8363547,combalia2019bcn20000, mitsuhara2019embedding}, a dermatologic dataset designed for skin cancer detection, featuring images of both benign and malignant lesions. Using the Reveal2Revise framework~\cite{pahde_reveal_2023}, we identify three artifacts naturally occurring in the dataset, strongly correlating with class labels:  ``band-aid'' (correlating with ``NV''), ``skin marker'' (``NV'', ``BKL''), and ``reflection'' (``BKL'') artifacts. To evaluate our approach, we additionally create a \emph{poisoned} test set for ISIC2019, where all samples have a ``reflection'' artifact superimposed on them. Visually the ``reflection'' artifact simulates the spot-wise reflection of a bright illumination source. Additional details for all the datasets are available in \cref{datasets}.

For all settings, we train  VGG16~\cite{simonyan2014very}, ResNet18~\cite{he_deep_2016}, and EfficientNet-B0~\cite{pmlr-v97-tan19a} models, with detailed training information provided in \cref{train}.

\subsection{\gls{cav} Calculation} \label{cav_calculation}
\begin{figure}[t]
  \includegraphics[width=\columnwidth]{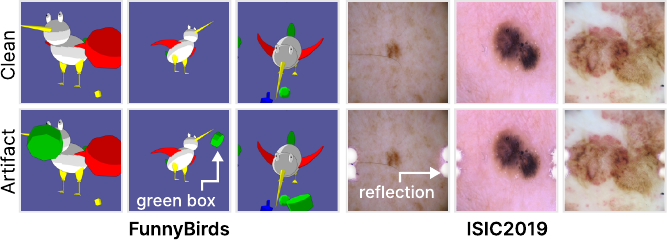}
  \centering
  \caption{Examples of adding artifact concepts in a controlled manner: (\emph{Left}): For FunnyBirds, we insert a ``green box'' into images. (\emph{Right}): For ISIC, we insert ``reflections'' on the side.}
\label{fig:generated}
\end{figure}

For each experimental setting, we generate datasets of the same images with and without artifacts. For both FunnyBirds datasets and each artifact, we generate pairs of images with and without the artifact. For the ISIC dataset and the ``reflection'' artifact, we superimpose examples of the artifact onto clean images from the associated class (refer to \cref{datasets} for details). \cref{fig:generated} displays some examples from the generated sets.

This process allows us to compute precise \gls{cav}s using sets of positive and negative samples. We refer to \gls{cav}s obtained through this method as ``generated''. As compared to the conventional method of calculating \gls{cav}s using the subsets from the dataset, we expect that the generated \gls{cav}s will provide more precise descriptions of the concept direction. Furthermore, the generated sets allow us to calculate pairwise concept directions. By comparing concept directions with \gls{cav}s, we assess the faithfulness of these concept representations in \cref{concepts_related}.

While each pair of images only differs in the presence or absence of an artifact, the difference between the activation vectors of the two samples may still not accurately represent the true concept direction. For instance, as depicted in the first example in \cref{fig:generated}, adding a ``green box'' artifact significantly obscures the beak concept. Therefore, we can expect that the activation vectors of the two samples also differ in their expression of the beak concept. However, for \gls{cav} calculation, we use large sets of pairs, which helps mitigate this effect. Further details about the \gls{cav} computation procedure can be found in \cref{cav_details}.

\begin{table}[t]
    \centering
    \small
    \begin{tabularx}{\linewidth}{p{0.10\textwidth}YYYY}
        \toprule
        \multirow{2}{*}{\gls{cav} Dataset} & \multicolumn{2}{c}{\specialcell{FunnyBirds \\ (``green box'')}} & \multicolumn{2}{c}{\specialcell{ISIC2019 \\ (``reflection'')}} \\

         & \multicolumn{1}{c}{Pattern} & \multicolumn{1}{c}{Filter} & \multicolumn{1}{c}{Pattern} & \multicolumn{1}{c}{Filter} \\
        \midrule
        Generated & $\textbf{0.824}$ & $0.101$ & $\textbf{0.617}$ & $0.343$ \\
        Data Subset & $0.563$ & $0.042$ & $0.469$ & $0.328$ \\
        \bottomrule
    \end{tabularx}
    \caption{Evaluating the alignment of computed artifact \glspl{cav} in terms of cosine similarity with the actual change in activations when the concept is added in controlled fashion in ResNet18 models. We compare \glspl{cav} computed on original data subsets, and pairs of clean and (generated) poisoned samples.}
    \label{tab:cav_alignment}
\end{table}

\subsection{Orthogonality of Concept Representations} \label{concepts_related}

In \cref{reactive_framework}, we introduced the reactive framework for model correction. Regardless of how precise and disentangled a concept representation used by a model correction method may be, the reactive approach can prove beneficial, \eg in cases when an artifact irrelevant for a certain class represents a useful feature for another (\eg, ``stripes''). However, many model correction methods may also face challenges related to the entanglement of concept representations. In this context, the reactive approach also enables us to ``minimize the damage'' on non-artifact samples.

In the following experiments, we evaluate the assumption of concept orthogonality in two controlled settings. Firstly, we investigate the concept representation of the ``green box'' artifact in the ResNet18 model trained on the backdoored FunnyBirds dataset. Secondly, we analyze the concept representation of the ``reflection'' artifact in the ResNet18 model trained on ISIC2019. Specifically, we use \gls{cav}s as concept representations since they are utilized in both \gls{pclarc} and our proposed approach \gls{rclarc}.

Initially, we assess the faithfulness of the \gls{cav} concept representation by comparing them to the pairwise concept directions. For both settings, we compute filter- and pattern-based \gls{cav}s using both the generated sets and sampled dataset examples. Then, we determine the alignment score by averaging the cosine similarity of each concept direction and \gls{cav}, following the approach in~\cite{pahde_navigating_2024}. The results are presented in \cref{tab:cav_alignment}. Confirming the findings of~\cite{pahde_navigating_2024}, pattern-based \gls{cav} performed better in both settings. Additionally, as anticipated, the generated \gls{cav}s offer a more accurate concept representation. We utilize the pattern-based \gls{cav} calculated on the generated sets for our subsequent experiments. Additional details regarding the evaluation of \gls{cav}s can be found in \cref{cav_eval}.

In our experiment investigating the orthogonality hypothesis, we analyze the distribution of the \gls{cav} activations of 500 randomly selected clean samples and all samples containing the artifact. Based on an assumption of the \gls{pclarc} framework, there should be no variance along the artifact direction for all clean samples. 

\begin{figure}[t]
  \includegraphics[width=\columnwidth]{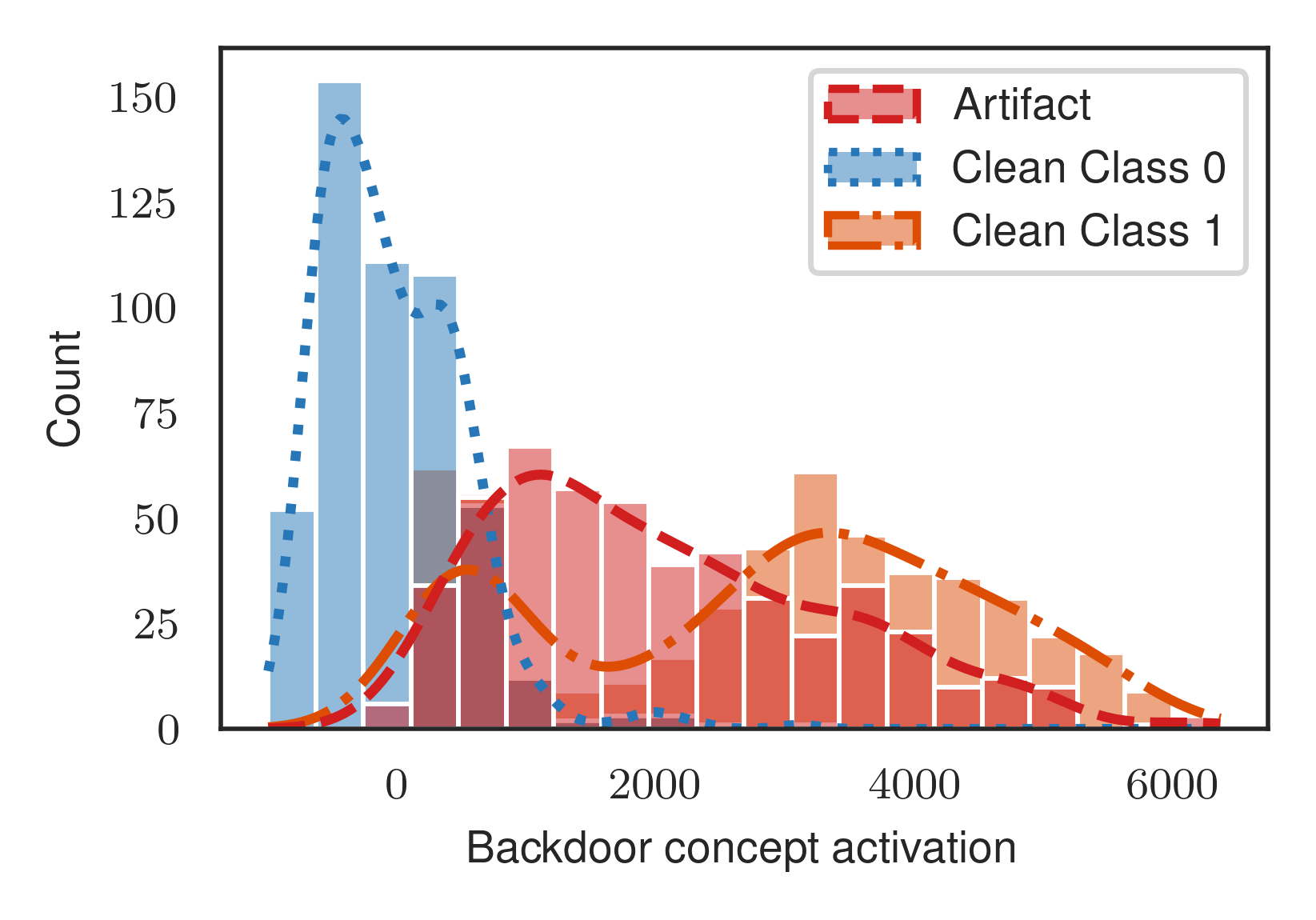}
  \centering
  \caption{Histogram of \gls{cav} activations for the FunnyBirds backdoor (``green box'') artifact in ResNet18:  the backdoor \gls{cav} aligns with features specific for class 1.}
\label{fig:fb_density}
\end{figure}

From \cref{fig:fb_density}, we can observe the distribution of pattern-based \gls{cav} activations for the backdoor FunnyBirds model. Clean samples exhibit considerable variance along the \gls{cav} direction. Specifically, samples belonging to class 1 notably activate the backdoor \gls{cav}, indicating entanglement between the backdoor concept and the concept of the class label which is predicted in the backdoor presence. Consequently, suppressing the artifact direction in this scenario coincidentally shifts the clean samples of class 1 towards or across the decision boundary towards class 0.

\begin{figure}[t]
  \includegraphics[width=\columnwidth]{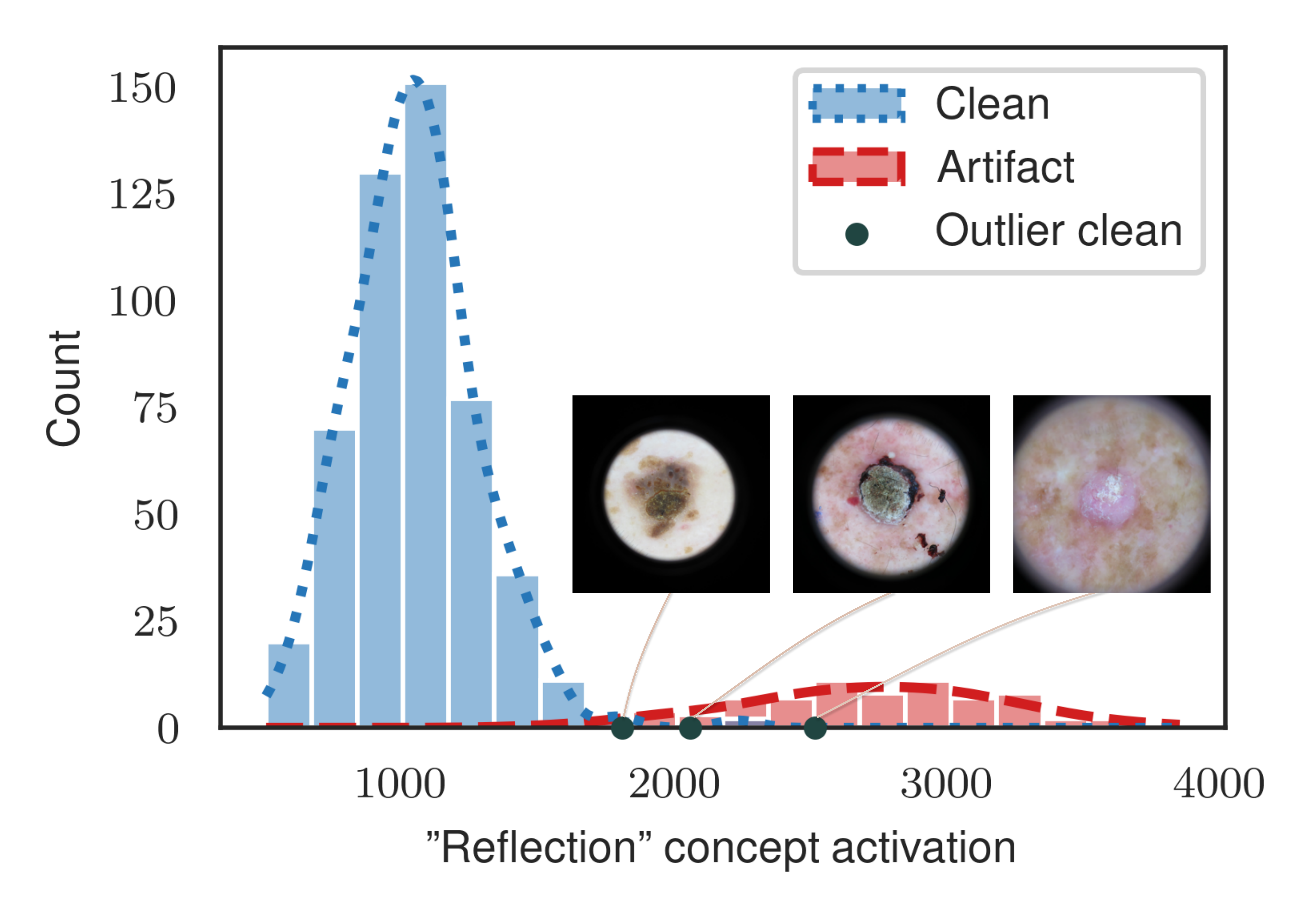}
  \centering
  \caption{Histogram of \gls{cav} activations for the ISIC ``reflection'' artifact in ResNet18: outlier clean samples with white spots lead to high concept activation.}
\label{fig:isic_density}
\end{figure}

\begin{figure}[ht]
  \includegraphics[width=\columnwidth]{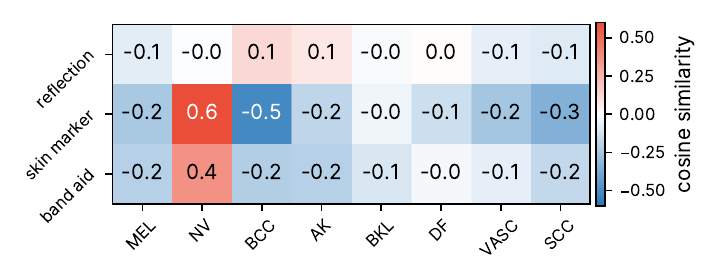}
  \centering
  \caption{Cosine similarity between artifact \gls{cav} and mean feature direction of each class for the ISIC dataset and ResNet-18: the artifact concept representations are entangled with clean features.}
\label{fig:isic_class_ortho}
\end{figure}

\cref{fig:isic_density} illustrates the distribution of clean and ``reflection''-artifact ISIC samples along the \gls{cav} direction. Firstly, clean samples exhibit variance along the artifact \gls{cav}, indicating that the direction encodes information unrelated to the artifact. Secondly, we observe a high level of activation of the concept in samples displaying features related to the ``reflection'' artifact, such as pale skin or white skin patches. These features could be important for prediction, suggesting that suppressing this information may not be desirable. \cref{fig:isic_class_ortho} further illustrates the cosine similarities between the \gls{cav} directions of the three ISIC artifacts and the class directions of clean samples. The non-orthogonality between the \gls{cav} and the class directions suggests that suppressing the artifact directions could impact the predictions of clean samples.

\begin{table*}[t]
    \centering
    \small
    \begin{tabularx}{\textwidth}{p{0.05\textwidth}p{0.08\textwidth}p{0.08\textwidth}AAAAA}
\toprule 
        \multicolumn{1}{l}{Model} & \multicolumn{1}{l}{Method}  & \multicolumn{1}{l}{Condition}& \multicolumn{2}{c}{\centering \specialcell{Accuracy \\ (clean)}} & \multicolumn{2}{c}{\centering \specialcell{Accuracy \\ (artifact)}} & \multicolumn{2}{c}{\centering \specialcell{F1 \\ (clean)}} & \multicolumn{2}{c}{\centering \specialcell{F1 \\ (artifact)}} & \multicolumn{2}{c}{\centering \specialcell{Artifact \\ relevance}} \\
        
        \midrule
\multirow[c]{5}{*}{\rotatebox{90}{Resnet18}} & Vanilla & - & 98.8 \text{ } &\vert  \text{ }93.3 & 12.1 \text{ } &\vert  \text{ }100.0 & 98.7 \text{ } &\vert  \text{ }92.4 & 10.8 \text{ } &\vert  \text{ }100.0 & 28.6 \text{ } &\vert  \text{ }5.2 \\
\cmidrule{2-13}
 & \gls{pclarc} & - & 40.1 \text{ } &\vert  \text{ }3.6 & 100.0 \text{ } &\vert  \text{ }100.0 & 28.6 \text{ } &\vert  \text{ }0.7 & 100.0 \text{ } &\vert  \text{ }100.0 & \textbf{17.9} \text{ } &\vert  \text{ }\textbf{3.8} \\
 & \multirow[t]{3}{*}{\gls{rclarc}} & Class & 40.1 \text{ } &\vert  \text{ }\textbf{93.3} & 100.0 \text{ } &\vert  \text{ }100.0 & 28.6 \text{ } &\vert  \text{ }\textbf{92.4} & 100.0 \text{ } &\vert  \text{ }100.0 & 18.4 \text{ } &\vert  \text{ }\textbf{3.8} \\
 &  & Artifact & \textbf{97.6} \text{ } &\vert  \text{ }92.6 & 97.0 \text{ } &\vert  \text{ }100.0 & \textbf{97.5} \text{ } &\vert  \text{ }91.6 & 49.2 \text{ } &\vert  \text{ }100.0 & 18.0 \text{ } &\vert  \text{ }\textbf{3.8} \\
 &  & Both & \textbf{97.6} \text{ } &\vert  \text{ }\textbf{93.3} & 97.0 \text{ } &\vert  \text{ }100.0 & \textbf{97.5} \text{ } &\vert  \text{ }\textbf{92.4} & 49.2 \text{ } &\vert  \text{ }100.0 & 18.4 \text{ } &\vert  \text{ }\textbf{3.8} \\
\midrule
\multirow[c]{5}{*}{\rotatebox{90}{VGG16}} & Vanilla & - & 98.8 \text{ } &\vert  \text{ }91.8 & 27.3 \text{ } &\vert  \text{ }97.0 & 98.7 \text{ } &\vert  \text{ }91.4 & 21.4 \text{ } &\vert  \text{ }32.8 & 22.7 \text{ } &\vert  \text{ }5.2 \\
\cmidrule{2-13}
 & \gls{pclarc} & - & 59.9 \text{ } &\vert  \text{ }4.6 & 0.0 \text{ } &\vert  \text{ }100.0 & 37.5 \text{ } &\vert  \text{ }2.4 & 0.0 \text{ } &\vert  \text{ }100.0 & 11.8 \text{ } &\vert  \text{ }4.5 \\
 & \multirow[t]{3}{*}{\gls{rclarc}} & Class & \textbf{98.8} \text{ } &\vert  \text{ }\textbf{91.8} & 27.3 \text{ } &\vert  \text{ }97.0 & \textbf{98.7} \text{ } &\vert  \text{ }\textbf{91.4} & 21.4 \text{ } &\vert  \text{ }32.8 & 11.3 \text{ } &\vert  \text{ }\textbf{4.4} \\
 &  & Artifact & 95.2 \text{ } &\vert  \text{ }87.6 & 3.0 \text{ } &\vert  \text{ }100.0 & 94.9 \text{ } &\vert  \text{ }85.9 & 2.9 \text{ } &\vert  \text{ }100.0 & \textbf{9.9} \text{ } &\vert  \text{ }\textbf{4.4} \\
 &  & Both & \textbf{98.8} \text{ } &\vert  \text{ }\textbf{91.8} & 27.3 \text{ } &\vert  \text{ }97.0 & \textbf{98.7} \text{ } &\vert  \text{ }\textbf{91.4} & 21.4 \text{ } &\vert  \text{ }32.8 & 11.3 \text{ } &\vert  \text{ }\textbf{4.4} \\
\midrule
\multirow[c]{5}{*}{\rotatebox{90}{Efficient Net-B0}} & Vanilla & - & 99.4 \text{ } &\vert  \text{ }86.6 & 6.1 \text{ } &\vert  \text{ }97.0 & 99.4 \text{ } &\vert  \text{ }83.1 & 5.7 \text{ } &\vert  \text{ }32.8 & 37.0 \text{ } &\vert  \text{ }4.0 \\
\cmidrule{2-13}
 & \gls{pclarc} & - & 91.6 \text{ } &\vert  \text{ }10.7 & 3.0 \text{ } &\vert  \text{ }0.0 & 90.9 \text{ } &\vert  \text{ }1.9 & 2.9 \text{ } &\vert  \text{ }0.0 & \textbf{24.8} \text{ } &\vert  \text{ }\textbf{3.5} \\
 & \multirow[t]{3}{*}{\gls{rclarc}} & Class & \textbf{99.4} \text{ } &\vert  \text{ }\textbf{83.4} & 6.1 \text{ } &\vert  \text{ }0.0 & \textbf{99.4} \text{ } &\vert  \text{ }\textbf{74.3} & 5.7 \text{ } &\vert  \text{ }0.0 & 24.9 \text{ } &\vert  \text{ }\textbf{3.5} \\
 &  & Artifact & \textbf{99.4} \text{ } &\vert  \text{ }82.9 & 3.0 \text{ } &\vert  \text{ }1.5 & \textbf{99.4} \text{ } &\vert  \text{ }73.9 & 2.9 \text{ } &\vert  \text{ }1.5 & \textbf{24.8} \text{ } &\vert  \text{ }\textbf{3.5} \\
 &  & Both & \textbf{99.4} \text{ } &\vert  \text{ }\textbf{83.4} & 6.1 \text{ } &\vert  \text{ }1.5 & \textbf{99.4} \text{ } &\vert  \text{ }\textbf{74.3} & 5.7 \text{ } &\vert  \text{ }1.0 & 24.9 \text{ } &\vert  \text{ }\textbf{3.5} \\
\bottomrule
    \end{tabularx}
        \caption{Model correction results for the background object artifacts inserted in FunnyBirds datasets. We report scores on the (\emph{backdoor} $|$ \emph{shortcuts}) version. The best scores are highlighted in bold.}
    \label{tab:EVAL_FUNNYBIRDS}
\end{table*}

\begin{table*}[t]
    \centering
    \small
    \begin{tabularx}{\textwidth}{p{0.05\textwidth}p{0.08\textwidth}p{0.08\textwidth}AAAAA}
\toprule 
        \multicolumn{1}{l}{Model} & \multicolumn{1}{l}{Method}  & \multicolumn{1}{l}{Condition}& \multicolumn{2}{c}{\centering \specialcell{Accuracy \\ (clean)}} & \multicolumn{2}{c}{\centering \specialcell{Accuracy \\ (artifact)}} & \multicolumn{2}{c}{\centering \specialcell{F1 \\ (clean)}} & \multicolumn{2}{c}{\centering \specialcell{F1 \\ (artifact)}} & \multicolumn{2}{c}{\centering \specialcell{Artifact \\ relevance}} \\
        
        \midrule
\multirow[c]{5}{*}{\rotatebox{90}{Resnet18}} & Vanilla & - & - \text{ } &\vert  \text{ }83.3 & 50.1 \text{ } &\vert  \text{ }80.4 & - \text{ } &\vert  \text{ }79.4 & 50.2 \text{ } &\vert  \text{ }33.4 & 23.9 \text{ } &\vert  \text{ }9.1 \\
\cmidrule{2-13}
 & \gls{pclarc} & - & - \text{ } &\vert  \text{ }61.0 & 45.1 \text{ } &\vert  \text{ }80.4 & - \text{ } &\vert  \text{ }21.5 & 14.4 \text{ } &\vert  \text{ }40.2 & \textbf{19.0} \text{ } &\vert  \text{ }\textbf{8.2} \\
 & \multirow[t]{3}{*}{\gls{rclarc}} & Class & - \text{ } &\vert  \text{ }\textbf{83.3} & \textbf{59.5} \text{ } &\vert  \text{ }80.4 & - \text{ } &\vert  \text{ }\textbf{79.1} & \textbf{52.4} \text{ } &\vert  \text{ }32.9 & 19.1 \text{ } &\vert  \text{ }8.3 \\
 &  & Artifact & - \text{ } &\vert  \text{ }83.0 & 46.2 \text{ } &\vert  \text{ }82.1 & - \text{ } &\vert  \text{ }79.0 & 43.8 \text{ } &\vert  \text{ }41.1 & 22.1 \text{ } &\vert  \text{ }8.6 \\
 &  & Both & - \text{ } &\vert  \text{ }83.0 & 50.3 \text{ } &\vert  \text{ }80.4 & - \text{ } &\vert  \text{ }79.0 & 50.2 \text{ } &\vert  \text{ }32.9 & 21.9 \text{ } &\vert  \text{ }8.6 \\
\midrule
\multirow[c]{5}{*}{\rotatebox{90}{VGG16}} & Vanilla & - & - \text{ } &\vert  \text{ }79.5 & 20.1 \text{ } &\vert  \text{ }78.6 & - \text{ } &\vert  \text{ }73.2 & 17.7 \text{ } &\vert  \text{ }32.4 & 46.3 \text{ } &\vert  \text{ }18.3 \\
\cmidrule{2-13}
 & \gls{pclarc} & - & - \text{ } &\vert  \text{ }63.8 & \textbf{56.8} \text{ } &\vert  \text{ }76.8 & - \text{ } &\vert  \text{ }30.9 & 24.3 \text{ } &\vert  \text{ }39.3 & \textbf{21.1} \text{ } &\vert  \text{ }15.1 \\
 & \multirow[t]{3}{*}{\gls{rclarc}} & Class & - \text{ } &\vert  \text{ }78.2 & 47.7 \text{ } &\vert  \text{ }78.6 & - \text{ } &\vert  \text{ }72.4 & \textbf{32.1} \text{ } &\vert  \text{ }32.4 & 27.2 \text{ } &\vert  \text{ }\textbf{14.5} \\
 &  & Artifact & - \text{ } &\vert  \text{ }\textbf{79.2} & 27.6 \text{ } &\vert  \text{ }82.1 & - \text{ } &\vert  \text{ }72.4 & 15.4 \text{ } &\vert  \text{ }41.6 & 38.6 \text{ } &\vert  \text{ }17.0 \\
 &  & Both & - \text{ } &\vert  \text{ }\textbf{79.2} & 27.4 \text{ } &\vert  \text{ }83.9 & - \text{ } &\vert  \text{ }\textbf{73.0} & 20.8 \text{ } &\vert  \text{ }34.4 & 38.7 \text{ } &\vert  \text{ }17.2 \\
\midrule
\multirow[c]{5}{*}{\rotatebox{90}{Efficient Net-B0}} & Vanilla & - & - \text{ } &\vert  \text{ }85.3 & 57.9 \text{ } &\vert  \text{ }83.9 & - \text{ } &\vert  \text{ }82.0 & 57.1 \text{ } &\vert  \text{ }35.0 & 22.2 \text{ } &\vert  \text{ }11.2 \\
\cmidrule{2-13}
 & \gls{pclarc} & - & - \text{ } &\vert  \text{ }51.5 & 51.7 \text{ } &\vert  \text{ }60.7 & - \text{ } &\vert  \text{ }8.5 & 8.5 \text{ } &\vert  \text{ }18.9 & 19.4 \text{ } &\vert  \text{ }9.9 \\
 & \multirow[t]{3}{*}{\gls{rclarc}} & Class & - \text{ } &\vert  \text{ }78.6 & \textbf{65.7} \text{ } &\vert  \text{ }58.9 & - \text{ } &\vert  \text{ }71.6 & 53.0 \text{ } &\vert  \text{ }15.0 & 18.5 \text{ } &\vert  \text{ }\textbf{9.5} \\
 &  & Artifact & - \text{ } &\vert  \text{ }83.5 & 59.8 \text{ } &\vert  \text{ }60.7 & - \text{ } &\vert  \text{ }79.8 & 45.1 \text{ } &\vert  \text{ }18.9 & 18.4 \text{ } &\vert  \text{ }9.7 \\
 &  & Both & - \text{ } &\vert  \text{ }\textbf{84.4} & 63.0 \text{ } &\vert  \text{ }58.9 & - \text{ } &\vert  \text{ }\textbf{81.2} & \textbf{56.3} \text{ } &\vert  \text{ }15.0 & \textbf{17.7} \text{ } &\vert  \text{ }\textbf{9.5} \\
\bottomrule
    \end{tabularx}
        \caption{Model correction results for the ``reflection'', ``band aid'' and ``skin marker'' artifacts found in ISIC. We report scores on the (\emph{poisoned} $|$ \emph{original}) version. The best scores are highlighted in bold.}
    \label{tab:EVAL_ISIC}
\end{table*}

\subsection{Model Correction with \gls{rclarc}} \label{performance}

In the following analysis, we examine the impact of incorporating reactivity into the \gls{pclarc} model correction method, based on three metrics: accuracy, F1 score, and artifact relevance, as measured by the share of Layer-wise Relevance Propagation (LRP)~\cite{10.1371/journal.pone.0130140} attribution in the artifact region. Our evaluation includes \gls{rclarc} with class- and artifact-condition-generating functions, as well as their combination. In all settings except for the poisoned ISIC2019, our primary focus is on evaluating the performance on clean samples. We expect the reactive approach to improve the preservation of performance on clean samples while still reducing artifact relevance. In the poisoned ISIC2019 setting, where clean samples are not present, we are mainly interested in recovering performance compared to the original Vanilla model. Additionally, we provide accuracy and F1 scores for artifact samples. Additional details regarding the ClArC parameters and the evaluation procedure are outlined in \cref{correction}.

\begin{figure}[ht]
  \includegraphics[width=\columnwidth]{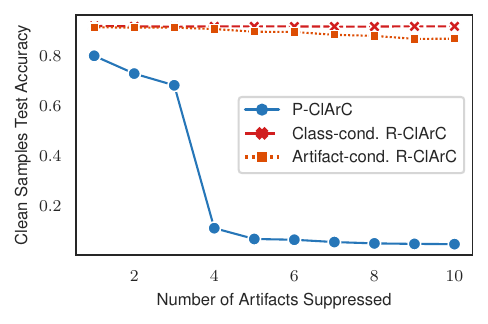}
  \centering
  \caption{Clean samples test accuracy by the number of artifacts suppressed with FunnyBirds \emph{shortcut} dataset and VGG16: with \emph{reactive} model correction (\gls{rclarc}), we can retain accuracy significantly better than when with the traditional approach (\gls{pclarc}).}
\label{fig:n_artifacts}
\end{figure}

The outcomes of model correction for VGG16, ResNet18, and EfficientNet-B0 on the FunnyBirds datasets are presented in \cref{tab:EVAL_FUNNYBIRDS}. In the \emph{backdoor} setting, the \gls{rclarc} class condition is met for the prediction of class 1 and in the \emph{shortcut} setting for the prediction of class 0. As expected, the \emph{backdoor} Vanilla models exhibit low performance on the backdoor samples. 
As mentioned in \cref{concepts_related}, in ResNet18, the artifact and class 1 concepts are correlated. Suppressing the artifact often leads to predictions of class 0 for class 1 samples, resulting in less than 50\% accuracy for \gls{pclarc} and class-conditional \gls{rclarc}. However, using artifact-conditional and combined approaches helps to reduce the backdoor effect while maintaining good performance on clean samples. Although \gls{pclarc} fails to reduce the backdoor effect for VGG16 and EfficientNet-B0, all variants of \gls{rclarc} allow us to recover accuracy and F1 scores for clean samples. In the \emph{shortcut} setting, suppressing 10 artifacts with \gls{pclarc} notably decreases performance. \cref{fig:n_artifacts} further illustrates the dynamics of clean sample accuracy as the number of corrected artifacts increases for the VGG16 model. All reactive ClArCs improve clean sample performance, with the combined approach yielding the best results across all architectures.

\begin{figure}[ht]
  \includegraphics[width=\columnwidth]{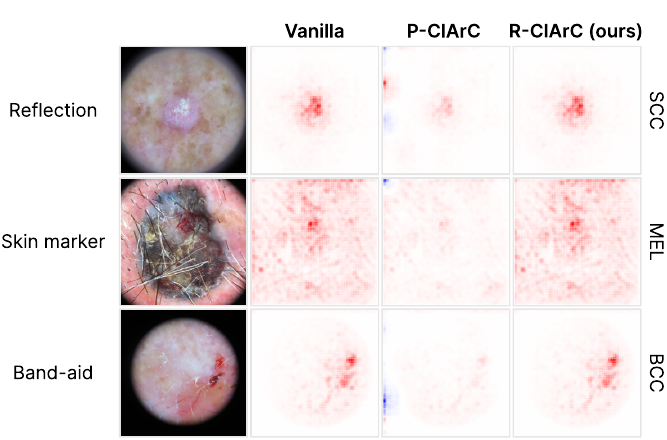}
  \centering
  \caption{LRP heatmaps for clean images strongly activating artifact concepts ``reflection'', ``band-aid,'' and ``skin marker'': \gls{pclarc} strongly suppresses \emph{all} artifacts, while \gls{rclarc} (combined approach) correctly refrains from suppression.}
\label{fig:heatmaps}
\end{figure}

\cref{tab:EVAL_ISIC} showcases the results for the same architectures, the ISIC2019 dataset and the artifacts ``band-aid,'' ``skin marker,'' and ``reflection'' for both the \emph{poisoned} (left) and \emph{real} (right) evaluation settings. In both configurations, the \gls{rclarc} class conditions link the class ``NV'' with artifacts ``band-aid'' and ``skin marker'', and the class ``BKL'' with ``skin marker'' and ``reflection''. In the \emph{poisoned} setting, we observe that \gls{rclarc} consistently outperforms Vanilla models across all architectures, which is not the case with \gls{pclarc} for ResNet18 and EfficientNet-B0. As all test samples are poisoned in this setting, global suppression of the artifact with \gls{pclarc} may prove beneficial, as indicated by the superior performance of \gls{pclarc} compared to \gls{rclarc} in the VGG16 case. In the \emph{real} setting, while \gls{pclarc} significantly harms the performance on clean samples, \gls{rclarc} recovers it to a level close to that of the Vanilla models for all architectures. \cref{fig:heatmaps} additionally depicts LRP heatmaps for the Vanilla models, as well as models corrected with \gls{pclarc} and \gls{rclarc}.

The condition-generating functions may not always be accurate (as demonstrated in \cref{correction}), causing the correction not to get activated when needed. This explains why in some cases \gls{pclarc} demonstrates slightly lower artifact relevance compared to \gls{rclarc}, and in the case of \emph{poisoned} ISIC2019 and VGG16 better model performance.
\section{Limitations}

While our results show that \gls{rclarc} outperforms \gls{pclarc} in various scenarios, the choice of reactive heuristics is not always straightforward. Furthermore, condition-generating functions may not always be accurate, resulting in the correction not being activated when required. 

Additionally, while reactive bias suppression reduces potentially negative (harmful) impact for clean samples (without present artifacts), it does not fully address all the flaws of \gls{pclarc}. 
Applying \gls{pclarc} transformations only on artifact samples may still inadvertently alter valid concepts crucial for prediction. Furthermore, merely suppressing a single direction in the latent space may not always suffice to eliminate artifact relevance, as evidenced in \cref{performance}.
\section{Conclusion}

In this work, we introduced a general framework for reactive model correction, aimed at mitigating the issue of ``collateral damage'' to task-relevant features. Specifically, we applied the reactive principle to the \gls{pclarc} post-hoc model correction framework, resulting in its reactive variant, \gls{rclarc}. Our empirical findings demonstrate the detrimental impact of \gls{pclarc} on clean samples while showcasing the superior performance of \gls{rclarc} in this aspect.

Future directions include application of the reactive principle to other post-hoc correction methods, such as classifier editing~\cite{santurkar_editing_2021}, LEACE\cite{belrose2023leace}, and INLP~\cite{ravfogel-etal-2020-null}. We also aim to explore more granular reactive conditions, such as those based on prediction strategy or artifact relevance.

\subsection*{Acknowledgements}
This work was supported by the Federal Ministry of Education and Research (BMBF) as grant BIFOLD (01IS18025A, 01IS180371I); 
the European Union’s Horizon 2020 research and innovation programme (EU Horizon 2020) as grant iToBoS (965221);
the European Union’s Horizon Europe research and innovation programme (EU Horizon Europe) as grant TEMA (101093003);
the German Research Foundation (DFG) as research unit DeSBi (KI-FOR 5363);   and the state of Berlin within the innovation support programme ProFIT (IBB) as grant BerDiBa (10174498).

{
    \small
    \bibliographystyle{ieeenat_fullname}
    \bibliography{main}

\begin{thebibliography}{44}
\providecommand{\natexlab}[1]{#1}
\providecommand{\url}[1]{\texttt{#1}}
\expandafter\ifx\csname urlstyle\endcsname\relax
  \providecommand{\doi}[1]{doi: #1}\else
  \providecommand{\doi}{doi: \begingroup \urlstyle{rm}\Url}\fi

\bibitem[Achtibat et~al.(2023)Achtibat, Dreyer, Eisenbraun, Bosse, Wiegand, Samek, and Lapuschkin]{achtibat_attribution_2023}
Reduan Achtibat, Maximilian Dreyer, Ilona Eisenbraun, Sebastian Bosse, Thomas Wiegand, Wojciech Samek, and Sebastian Lapuschkin.
\newblock From attribution maps to human-understandable explanations through {Concept} {Relevance} {Propagation}.
\newblock \emph{Nature Machine Intelligence}, 5\penalty0 (9):\penalty0 1006--1019, 2023.
\newblock Number: 9 Publisher: Nature Publishing Group.

\bibitem[Anders et~al.(2021)Anders, Neumann, Samek, M{\"{u}}ller, and Lapuschkin]{anders2021zennit}
Christopher~J. Anders, David Neumann, Wojciech Samek, Klaus{-}Robert M{\"{u}}ller, and Sebastian Lapuschkin.
\newblock Software for dataset-wide {XAI:} from local explanations to global insights with zennit, corelay, and virelay.
\newblock \emph{CoRR}, abs/2106.13200, 2021.

\bibitem[Anders et~al.(2022)Anders, Weber, Neumann, Samek, Müller, and Lapuschkin]{anders_finding_2022}
Christopher~J. Anders, Leander Weber, David Neumann, Wojciech Samek, Klaus-Robert Müller, and Sebastian Lapuschkin.
\newblock Finding and removing {Clever} {Hans}: {Using} explanation methods to debug and improve deep models.
\newblock \emph{Information Fusion}, 77:\penalty0 261--295, 2022.

\bibitem[Bach et~al.(2015)Bach, Binder, Montavon, Klauschen, Müller, and Samek]{10.1371/journal.pone.0130140}
Sebastian Bach, Alexander Binder, Grégoire Montavon, Frederick Klauschen, Klaus-Robert Müller, and Wojciech Samek.
\newblock On pixel-wise explanations for non-linear classifier decisions by layer-wise relevance propagation.
\newblock \emph{PLOS ONE}, 10\penalty0 (7):\penalty0 1--46, 2015.

\bibitem[Belrose et~al.(2023)Belrose, Schneider-Joseph, Ravfogel, Cotterell, Raff, and Biderman]{belrose2023leace}
Nora Belrose, David Schneider-Joseph, Shauli Ravfogel, Ryan Cotterell, Edward Raff, and Stella Biderman.
\newblock Leace: Perfect linear concept erasure in closed form, 2023.

\bibitem[Bolukbasi et~al.(2016)Bolukbasi, Chang, Zou, Saligrama, and Kalai]{NIPS2016_a486cd07}
Tolga Bolukbasi, Kai-Wei Chang, James~Y Zou, Venkatesh Saligrama, and Adam~T Kalai.
\newblock Man is to computer programmer as woman is to homemaker? debiasing word embeddings.
\newblock In \emph{Advances in Neural Information Processing Systems}. Curran Associates, Inc., 2016.

\bibitem[Briganti and Le~Moine(2020)]{briganti2020artificial}
Giovanni Briganti and Olivier Le~Moine.
\newblock Artificial intelligence in medicine: today and tomorrow.
\newblock \emph{Frontiers in medicine}, 7:\penalty0 509744, 2020.

\bibitem[Codella et~al.(2018)Codella, Gutman, Celebi, Helba, Marchetti, Dusza, Kalloo, Liopyris, Mishra, Kittler, and Halpern]{8363547}
Noel C.~F. Codella, David Gutman, M.~Emre Celebi, Brian Helba, Michael~A. Marchetti, Stephen~W. Dusza, Aadi Kalloo, Konstantinos Liopyris, Nabin Mishra, Harald Kittler, and Allan Halpern.
\newblock Skin lesion analysis toward melanoma detection: A challenge at the 2017 international symposium on biomedical imaging (isbi), hosted by the international skin imaging collaboration (isic).
\newblock In \emph{2018 IEEE 15th International Symposium on Biomedical Imaging (ISBI 2018)}, pages 168--172, 2018.

\bibitem[Combalia et~al.(2019)Combalia, Codella, Rotemberg, Helba, Vilaplana, Reiter, Carrera, Barreiro, Halpern, Puig, and Malvehy]{combalia2019bcn20000}
Marc Combalia, Noel C.~F. Codella, Veronica Rotemberg, Brian Helba, Veronica Vilaplana, Ofer Reiter, Cristina Carrera, Alicia Barreiro, Allan~C. Halpern, Susana Puig, and Josep Malvehy.
\newblock Bcn20000: Dermoscopic lesions in the wild, 2019.

\bibitem[Cortes and Vapnik(1995)]{cortes_support-vector_1995}
Corinna Cortes and Vladimir Vapnik.
\newblock Support-vector networks.
\newblock \emph{Machine Learning}, 20\penalty0 (3):\penalty0 273--297, 1995.

\bibitem[De~Cao et~al.(2021)De~Cao, Aziz, and Titov]{de2021editing}
Nicola De~Cao, Wilker Aziz, and Ivan Titov.
\newblock Editing factual knowledge in language models.
\newblock In \emph{Proceedings of the 2021 Conference on Empirical Methods in Natural Language Processing}, pages 6491--6506, Online and Punta Cana, Dominican Republic, 2021. Association for Computational Linguistics.

\bibitem[Dodge et~al.(2021)Dodge, Sap, Marasovic, Agnew, Ilharco, Groeneveld, and Gardner]{dodge2021documenting}
Jesse Dodge, Maarten Sap, Ana Marasovic, William Agnew, Gabriel Ilharco, Dirk Groeneveld, and Matt Gardner.
\newblock Documenting the english colossal clean crawled corpus.
\newblock \emph{CoRR}, abs/2104.08758, 2021.

\bibitem[Dreyer et~al.(2023)Dreyer, Achtibat, Samek, and Lapuschkin]{dreyer_understanding_2023}
Maximilian Dreyer, Reduan Achtibat, Wojciech Samek, and Sebastian Lapuschkin.
\newblock Understanding the ({Extra}-){Ordinary}: {Validating} {Deep} {Model} {Decisions} with {Prototypical} {Concept}-based {Explanations}, 2023.
\newblock arXiv:2311.16681 [cs].

\bibitem[Dreyer et~al.(2024)Dreyer, Pahde, Anders, Samek, and Lapuschkin]{dreyer2023hope}
Maximilian Dreyer, Frederik Pahde, Christopher~J. Anders, Wojciech Samek, and Sebastian Lapuschkin.
\newblock From hope to safety: Unlearning biases of deep models via gradient penalization in latent space.
\newblock \emph{Proceedings of the AAAI Conference on Artificial Intelligence}, 38\penalty0 (19):\penalty0 21046--21054, 2024.

\bibitem[Fel et~al.(2023)Fel, Boutin, Béthune, Cadene, Moayeri, Andéol, Chalvidal, and Serre]{fel_holistic_2023}
Thomas Fel, Victor Boutin, Louis Béthune, Remi Cadene, Mazda Moayeri, Léo Andéol, Mathieu Chalvidal, and Thomas Serre.
\newblock A {Holistic} {Approach} to {Unifying} {Automatic} {Concept} {Extraction} and {Concept} {Importance} {Estimation}.
\newblock \emph{Advances in Neural Information Processing Systems}, 36, 2023.

\bibitem[Geirhos et~al.(2020)Geirhos, Jacobsen, Michaelis, Zemel, Brendel, Bethge, and Wichmann]{geirhos_shortcut_2020}
Robert Geirhos, Jörn-Henrik Jacobsen, Claudio Michaelis, Richard Zemel, Wieland Brendel, Matthias Bethge, and Felix~A. Wichmann.
\newblock Shortcut learning in deep neural networks.
\newblock \emph{Nature Machine Intelligence}, 2\penalty0 (11):\penalty0 665--673, 2020.
\newblock Number: 11 Publisher: Nature Publishing Group.

\bibitem[Gu et~al.(2019)Gu, Dolan-Gavitt, and Garg]{gu2019badnets}
Tianyu Gu, Brendan Dolan-Gavitt, and Siddharth Garg.
\newblock Badnets: Identifying vulnerabilities in the machine learning model supply chain, 2019.

\bibitem[He et~al.(2016)He, Zhang, Ren, and Sun]{he_deep_2016}
Kaiming He, Xiangyu Zhang, Shaoqing Ren, and Jian Sun.
\newblock Deep residual learning for image recognition.
\newblock In \emph{Proceedings of the IEEE Conference on Computer Vision and Pattern Recognition (CVPR)}, 2016.

\bibitem[Hesse et~al.(2023)Hesse, Schaub-Meyer, and Roth]{hesse_funnybirds_2023}
Robin Hesse, Simone Schaub-Meyer, and Stefan Roth.
\newblock Funnybirds: A synthetic vision dataset for a part-based analysis of explainable ai methods.
\newblock In \emph{Proceedings of the IEEE/CVF International Conference on Computer Vision (ICCV)}, pages 3981--3991, 2023.

\bibitem[Kim et~al.(2018)Kim, Wattenberg, Gilmer, Cai, Wexler, Viegas, and Sayres]{kim_interpretability_2018}
Been Kim, Martin Wattenberg, Justin Gilmer, Carrie Cai, James Wexler, Fernanda Viegas, and Rory Sayres.
\newblock Interpretability {Beyond} {Feature} {Attribution}: {Quantitative} {Testing} with {Concept} {Activation} {Vectors} ({TCAV}).
\newblock In \emph{Proceedings of the 35th {International} {Conference} on {Machine} {Learning}}, pages 2668--2677. PMLR, 2018.
\newblock ISSN: 2640-3498.

\bibitem[Kohlbrenner et~al.(2020)Kohlbrenner, Bauer, Nakajima, Binder, Samek, and Lapuschkin]{kohlbrenner2020towards}
Maximilian Kohlbrenner, Alexander Bauer, Shinichi Nakajima, Alexander Binder, Wojciech Samek, and Sebastian Lapuschkin.
\newblock Towards best practice in explaining neural network decisions with lrp.
\newblock In \emph{2020 International Joint Conference on Neural Networks (IJCNN)}, pages 1--7, 2020.

\bibitem[Kumar et~al.(2022)Kumar, Tan, and Sharma]{NEURIPS2022_725f5e80}
Abhinav Kumar, Chenhao Tan, and Amit Sharma.
\newblock Probing classifiers are unreliable for concept removal and detection.
\newblock In \emph{Advances in Neural Information Processing Systems}, pages 17994--18008. Curran Associates, Inc., 2022.

\bibitem[Lapuschkin et~al.(2019{\natexlab{a}})Lapuschkin, W{\"a}ldchen, Binder, Montavon, Samek, and M{\"u}ller]{lapuschkin2019unmasking}
Sebastian Lapuschkin, Stephan W{\"a}ldchen, Alexander Binder, Gr{\'e}goire Montavon, Wojciech Samek, and Klaus-Robert M{\"u}ller.
\newblock Unmasking clever hans predictors and assessing what machines really learn.
\newblock \emph{Nature Communications}, 10\penalty0 (1):\penalty0 1096, 2019{\natexlab{a}}.

\bibitem[Lapuschkin et~al.(2019{\natexlab{b}})Lapuschkin, Wäldchen, Binder, Montavon, Samek, and Müller]{lapuschkin_unmasking_2019}
Sebastian Lapuschkin, Stephan Wäldchen, Alexander Binder, Grégoire Montavon, Wojciech Samek, and Klaus-Robert Müller.
\newblock Unmasking {Clever} {Hans} predictors and assessing what machines really learn.
\newblock \emph{Nature Communications}, 10\penalty0 (1):\penalty0 1096, 2019{\natexlab{b}}.
\newblock Number: 1 Publisher: Nature Publishing Group.

\bibitem[Li et~al.(2023)Li, Evtimov, Gordo, Hazirbas, Hassner, Ferrer, Xu, and Ibrahim]{li2023whac}
Zhiheng Li, Ivan Evtimov, Albert Gordo, Caner Hazirbas, Tal Hassner, Cristian~Canton Ferrer, Chenliang Xu, and Mark Ibrahim.
\newblock A whac-a-mole dilemma: Shortcuts come in multiples where mitigating one amplifies others.
\newblock In \emph{Proceedings of the IEEE/CVF Conference on Computer Vision and Pattern Recognition}, pages 20071--20082, 2023.

\bibitem[Liang et~al.(2022)Liang, Tadesse, Ho, Fei-Fei, Zaharia, Zhang, and Zou]{liang2022advances}
Weixin Liang, Girmaw~Abebe Tadesse, Daniel Ho, Li Fei-Fei, Matei Zaharia, Ce Zhang, and James Zou.
\newblock Advances, challenges and opportunities in creating data for trustworthy ai.
\newblock \emph{Nature Machine Intelligence}, 4\penalty0 (8):\penalty0 669--677, 2022.

\bibitem[Linhardt et~al.(2024)Linhardt, Müller, and Montavon]{LINHARDT2024pruning}
Lorenz Linhardt, Klaus-Robert Müller, and Grégoire Montavon.
\newblock Preemptively pruning clever-hans strategies in deep neural networks.
\newblock \emph{Information Fusion}, 103:\penalty0 102094, 2024.

\bibitem[Mitsuhara et~al.(2019)Mitsuhara, Fukui, Sakashita, Ogata, Hirakawa, Yamashita, and Fujiyoshi]{mitsuhara2019embedding}
Masahiro Mitsuhara, Hiroshi Fukui, Yusuke Sakashita, Takanori Ogata, Tsubasa Hirakawa, Takayoshi Yamashita, and Hironobu Fujiyoshi.
\newblock Embedding human knowledge into deep neural network via attention map, 2019.

\bibitem[Neuhaus et~al.(2023)Neuhaus, Augustin, Boreiko, and Hein]{neuhaus2023spurious}
Yannic Neuhaus, Maximilian Augustin, Valentyn Boreiko, and Matthias Hein.
\newblock Spurious features everywhere-large-scale detection of harmful spurious features in imagenet.
\newblock In \emph{Proceedings of the IEEE/CVF International Conference on Computer Vision}, pages 20235--20246, 2023.

\bibitem[Pahde et~al.(2023)Pahde, Dreyer, Samek, and Lapuschkin]{pahde_reveal_2023}
Frederik Pahde, Maximilian Dreyer, Wojciech Samek, and Sebastian Lapuschkin.
\newblock Reveal to revise: An explainable ai life cycle for iterative bias correction of deep models.
\newblock In \emph{Medical Image Computing and Computer Assisted Intervention -- MICCAI 2023}, pages 596--606, Cham, 2023. Springer Nature Switzerland.

\bibitem[Pahde et~al.(2024)Pahde, Dreyer, Weber, Weckbecker, Anders, Wiegand, Samek, and Lapuschkin]{pahde_navigating_2024}
Frederik Pahde, Maximilian Dreyer, Leander Weber, Moritz Weckbecker, Christopher~J. Anders, Thomas Wiegand, Wojciech Samek, and Sebastian Lapuschkin.
\newblock Navigating {Neural} {Space}: {Revisiting} {Concept} {Activation} {Vectors} to {Overcome} {Directional} {Divergence}, 2024.
\newblock arXiv:2202.03482 [cs].

\bibitem[Ravfogel et~al.(2020)Ravfogel, Elazar, Gonen, Twiton, and Goldberg]{ravfogel-etal-2020-null}
Shauli Ravfogel, Yanai Elazar, Hila Gonen, Michael Twiton, and Yoav Goldberg.
\newblock Null it out: Guarding protected attributes by iterative nullspace projection.
\newblock In \emph{Proceedings of the 58th Annual Meeting of the Association for Computational Linguistics}, pages 7237--7256, Online, 2020. Association for Computational Linguistics.

\bibitem[Rieger et~al.(2020)Rieger, Singh, Murdoch, and Yu]{rieger2020interpretations}
Laura Rieger, Chandan Singh, William Murdoch, and Bin Yu.
\newblock Interpretations are useful: penalizing explanations to align neural networks with prior knowledge.
\newblock In \emph{International Conference on Machine Learning}, pages 8116--8126. PMLR, 2020.

\bibitem[Ross et~al.(2017)Ross, Hughes, and Doshi-Velez]{ross2017right}
Andrew~Slavin Ross, Michael~C Hughes, and Finale Doshi-Velez.
\newblock Right for the right reasons: training differentiable models by constraining their explanations.
\newblock In \emph{Proceedings of the 26th International Joint Conference on Artificial Intelligence}, pages 2662--2670, 2017.

\bibitem[Rouf et~al.(2021)Rouf, Malik, Arif, Sharma, Singh, Aich, and Kim]{rouf2021stock}
Nusrat Rouf, Majid~Bashir Malik, Tasleem Arif, Sparsh Sharma, Saurabh Singh, Satyabrata Aich, and Hee-Cheol Kim.
\newblock Stock market prediction using machine learning techniques: a decade survey on methodologies, recent developments, and future directions.
\newblock \emph{Electronics}, 10\penalty0 (21):\penalty0 2717, 2021.

\bibitem[Russakovsky et~al.(2015)Russakovsky, Deng, Su, Krause, Satheesh, Ma, Huang, Karpathy, Khosla, Bernstein, Berg, and Fei-Fei]{russakovsky_imagenet_2015}
Olga Russakovsky, Jia Deng, Hao Su, Jonathan Krause, Sanjeev Satheesh, Sean Ma, Zhiheng Huang, Andrej Karpathy, Aditya Khosla, Michael Bernstein, Alexander~C. Berg, and Li Fei-Fei.
\newblock {ImageNet} {Large} {Scale} {Visual} {Recognition} {Challenge}.
\newblock \emph{International Journal of Computer Vision}, 115\penalty0 (3):\penalty0 211--252, 2015.

\bibitem[Santurkar et~al.(2021)Santurkar, Tsipras, Elango, Bau, Torralba, and Madry]{santurkar_editing_2021}
Shibani Santurkar, Dimitris Tsipras, Mahalaxmi Elango, David Bau, Antonio Torralba, and Aleksander Madry.
\newblock Editing a classifier by rewriting its prediction rules.
\newblock In \emph{Advances in {Neural} {Information} {Processing} {Systems}}, pages 23359--23373. Curran Associates, Inc., 2021.

\bibitem[Schramowski et~al.(2020)Schramowski, Stammer, Teso, Brugger, Herbert, Shao, Luigs, Mahlein, and Kersting]{schramowski2020making}
Patrick Schramowski, Wolfgang Stammer, Stefano Teso, Anna Brugger, Franziska Herbert, Xiaoting Shao, Hans-Georg Luigs, Anne-Katrin Mahlein, and Kristian Kersting.
\newblock Making deep neural networks right for the right scientific reasons by interacting with their explanations.
\newblock \emph{Nature Machine Intelligence}, 2\penalty0 (8):\penalty0 476--486, 2020.

\bibitem[Simonyan and Zisserman(2015)]{simonyan2014very}
Karen Simonyan and Andrew Zisserman.
\newblock Very deep convolutional networks for large-scale image recognition.
\newblock \emph{International Conference on Learning Representations}, 2015.

\bibitem[Tan and Le(2019)]{pmlr-v97-tan19a}
Mingxing Tan and Quoc Le.
\newblock {E}fficient{N}et: Rethinking model scaling for convolutional neural networks.
\newblock In \emph{Proceedings of the 36th International Conference on Machine Learning}, pages 6105--6114. PMLR, 2019.

\bibitem[Teso and Kersting(2019)]{teso2019explanatory}
Stefano Teso and Kristian Kersting.
\newblock Explanatory interactive machine learning.
\newblock In \emph{Proceedings of the 2019 AAAI/ACM Conference on AI, Ethics, and Society}, pages 239--245, 2019.

\bibitem[Travaini et~al.(2022)Travaini, Pacchioni, Bellumore, Bosia, and De~Micco]{travaini2022machine}
Guido~Vittorio Travaini, Federico Pacchioni, Silvia Bellumore, Marta Bosia, and Francesco De~Micco.
\newblock Machine learning and criminal justice: A systematic review of advanced methodology for recidivism risk prediction.
\newblock \emph{International journal of environmental research and public health}, 19\penalty0 (17):\penalty0 10594, 2022.

\bibitem[Weng et~al.(2023)Weng, Pegios, Feragen, Petersen, and Bigdeli]{weng2023fast}
Nina Weng, Paraskevas Pegios, Aasa Feragen, Eike Petersen, and Siavash Bigdeli.
\newblock Fast diffusion-based counterfactuals for shortcut removal and generation.
\newblock \emph{arXiv preprint arXiv:2312.14223}, 2023.

\bibitem[Wu et~al.(2023)Wu, Yuksekgonul, Zhang, and Zou]{wu2023discover}
Shirley Wu, Mert Yuksekgonul, Linjun Zhang, and James Zou.
\newblock Discover and cure: Concept-aware mitigation of spurious correlation.
\newblock In \emph{International Conference on Machine Learning}, pages 37765--37786. PMLR, 2023.

\end{thebibliography}
}

\clearpage
\setcounter{page}{1}
\maketitlesupplementary
\appendix

\section{Method} \label{method_appx}

In the upcoming section, we explore further details and derivations of our methods. \cref{mult-cav-pclarc-derivation} provides the derivation of multi-artifact ~\gls{pclarc}, as per \cref{eq:mp-clarc}. \cref{pseudo} includes pseudocode for both \gls{pclarc} and \gls{rclarc}. Lastly, in \cref{3d}, we introduce a 3D toy experiment to illustrate the transformations induced by P- and \gls{rclarc}s.

\subsection{Derivation for multi-artifact \gls{pclarc}} \label{mult-cav-pclarc-derivation}

Let $\mathcal{C}' \subseteq \mathcal{C}$ represent a subset of artifacts with a size $|\mathcal{C}'| = k$, $1 \leq k \leq n$. Let $V_{\mathcal{C}'} =  \left[ v_i \right]_{c_i \in \mathcal{C}'}$ be the matrix comprised of the respective \gls{cav}s as column vectors. Let $Z_{\mathcal{C}'}^{-} = \bigcap_{{c_i \in \mathcal{C}'}} X^{-}_{i}$ be the intersection of negative examples, and $z_{\mathcal{C}'}^{-} = \frac{1}{|Z_{\mathcal{C}'}^{-}|} \sum_{z \in Z_{\mathcal{C}'}^{-}} a(z)$.
Let $a_x = a(x)$ and $p_x = \Tilde{h} (a_x, \mathcal{C}')$ be the \gls{pclarc} transformation. Our objective function can be then formalized as follows:

\begin{equation}
\begin{aligned}
\min_{p_x} || p_x - a_x||^2\\
\textrm{s.t. } V_{\mathcal{C}'}^T (p_x - z_{\mathcal{C}'}^{-}) = 0.\\
\end{aligned}
\end{equation}

The corresponding Lagrangian function is as follows:

\begin{equation}
\begin{aligned}
\mathcal{L} = \frac{1}{2}|| p_x - a_x||^2 + \lambda^T V_{\mathcal{C}'}^T (p_x - z_{\mathcal{C}'}^{-}).\\
\end{aligned}
\end{equation}

Applying the Karush-Kuhn-Tucker conditions, we get:

\begin{equation}
\begin{aligned}
p_x - a_x + V_{\mathcal{C}'} \lambda &= 0; \\
V_{\mathcal{C}'}^T (p_x - z_{\mathcal{C}'}^{-}) &= 0.\\
\end{aligned}
\end{equation}

Solving for $\lambda$ we get:

\begin{equation}
\begin{aligned}
V_{\mathcal{C}'}^Tp_x - V_{\mathcal{C}'}^T a_x + V_{\mathcal{C}'}^T V_{\mathcal{C}'} \lambda  = 0 &\iff \\
V_{\mathcal{C}'}^T z_{\mathcal{C}'}^{-} - V_{\mathcal{C}'}^T a_x + V_{\mathcal{C}'}^T V_{\mathcal{C}'} \lambda = 0 &\iff\\
\lambda = (V_{\mathcal{C}'}^T V_{\mathcal{C}'})^{-1}V_{\mathcal{C}'}^T (a_x - z_{\mathcal{C}'}^{-}).\\
\end{aligned}
\end{equation}

Inserting the resulting $\lambda$ into the first KKT condition we get:
\begin{equation}
\begin{aligned}
p_x - x + V_{\mathcal{C}'} (V_{\mathcal{C}'}^T V_{\mathcal{C}'})^{-1}V_{\mathcal{C}'}^T (a_x - z_{\mathcal{C}'}^{-}) = 0 &\iff \\
p_x = x - V_{\mathcal{C}'} (V_{\mathcal{C}'}^T V_{\mathcal{C}'})^{-1}V_{\mathcal{C}'}^T (a_x - z_{\mathcal{C}'}^{-}) &.
\end{aligned}
\end{equation}

Thus, we acquire $p_x = \Tilde{h}(x)$ as the \gls{pclarc} transformation for multiple artifacts, as described in \cref{eq:mp-clarc}.

\subsection{Pseudocode for \gls{pclarc} and \gls{rclarc}} \label{pseudo}
 A detailed Algorithm for both \gls{pclarc} and \gls{rclarc} shown in \cref{alg:clarc} under the common name of Class Artifact Compensation.

    \RestyleAlgo{algoruled}
        \begin{algorithm}[ht]
            \LinesNumbered
            \KwData{
                Sample $x$; \\
                Model $f$ with accessible layer $l$ (and subnetwork $f_l$);\\
                For each artifact in $\mathcal{C}$, sets of positive examples $X^{+} = \{X^{+}_1, X^{+}_2, \dots, X^{+}_n\}$ and negative examples $X^{-} = \{X^{-}_1, X^{-}_2, \dots, X^{-}_n\}$;\\
                For each artifact in $\mathcal{C}$, sets of activations of positive examples $A^{+} = \{A^{+}_1, A^{+}_2, \dots, A^{+}_n\}$ and activations of negative examples $A^{-} = \{A^{-}_1, A^{-}_2, \dots, A^{-}_n\}$ in  layer $l$;\\
                Set of layer-$l$ CAVs $V^l$ for each artifact in $\mathcal{C}$.\\
            }
            \KwResult{
                 output for $x$ according to a modified predictor $f'$ desensitized to artifacts $\mathcal{C}$
            }
            \tcc{deactivate the use of $\mathcal{C}$ in $f$}
            \uIf{\gls{pclarc}}
            {
                $Z^{-} =  \text{mean\_of\_intersection}(A^{-})$\;
                ${\color{rred} h_c^l } = \text{backward\_artifact\_model}(V^l, Z^{-})$\;
            
            }\uElseIf{\gls{rclarc}}
            {
            $\mathcal{C}' = \text{condition\_generating\_function}(x)$\;
            $V_{\mathcal{C}'}^l = \text{subset\_by\_concept}(V^l, \mathcal{C}')$\;
            $A_{\mathcal{C}'}^{-} = \text{subset\_by\_concept}(A^{-}, \mathcal{C}')$\;
            $Z_{\mathcal{C}'}^{-} =  \text{mean\_of\_intersection}(A_{\mathcal{C}'}^{-})$\;
            ${\color{rred} h_c^l } = \text{backward\_artifact\_model}(V_{\mathcal{C}'}^l, Z_{\mathcal{C}'}^{-})$\;

            }
            $a_x = f_L(x)$\;
            ${\color{red} f_{l'}}(a_{x}) \coloneqq h_c^l(a_{x})$\;
            $f' = f_L \circ \dots \circ f_{l+1} \circ {\color{red} f_{l'}} \circ f_l  \circ \dots \circ f_1(x)$\;
            \KwRet{$f'(x)$}
            
            \caption{\glsdesc{clarc}}
            \label{alg:clarc}
        \end{algorithm}

\begin{figure*}[ht]
  \includegraphics[width=0.9 \textwidth]{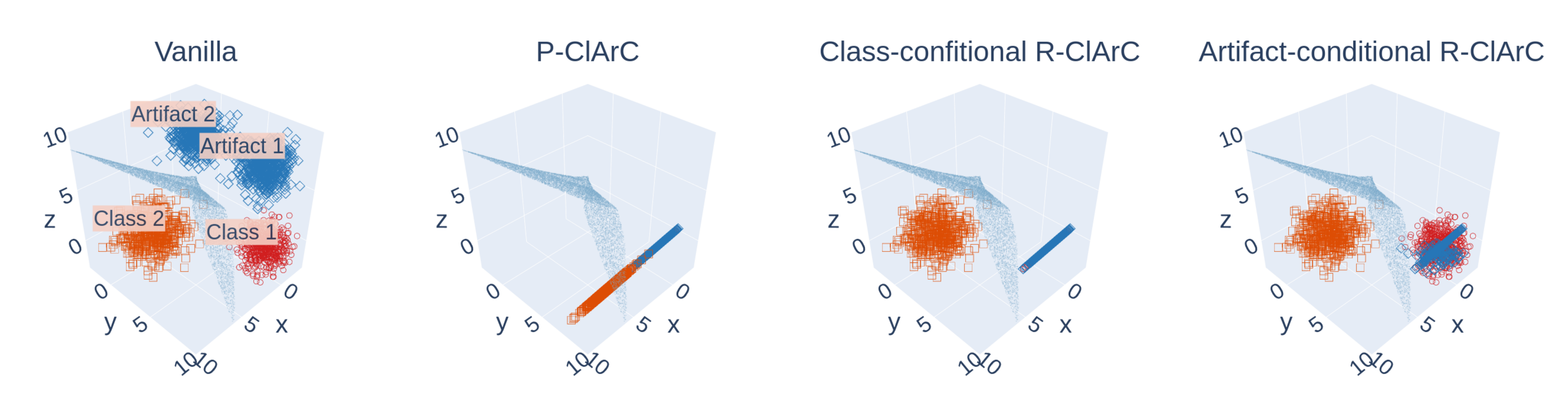}
  \centering
  \caption{3D Toy Model illustrating \gls{pclarc} and \gls{rclarc} transformations. The dataset includes Class 1, Class 2, and two artifacts belonging to Class 1. A three-layer feed-forward neural network is used for binary classification, with its decision boundary shown in light blue. \gls{pclarc} shifts Class 2 samples across the decision boundary, resulting in misclassification, while both versions of \gls{rclarc} maintain accuracy.}
\label{r-clarc-toy}
\end{figure*}

\subsection{3D Toy Model} \label{3d}

We construct a three-dimensional toy dataset comprising two classes. In Class 1, two artifacts are present. For Class 1 (red circles), we generate 500 clean samples distributed normally, with a mean at coordinates $(0, 8, 0)$ and a covariance matrix equal $I$, where $I$ represents the $3 \times 3$ identity matrix. Additionally, 500 samples are created for Artifact 1 (blue diamonds), centered at $(1, 8, 8)$, with covariance matrix $I$. Analogously, another set of 500 samples is generated for Artifact 2 (blue diamonds), with a mean at $(1, 1, 8)$ and covariance matrix $I$. For Class 2 (orange squares), 500 clean samples are distributed normally with a mean at $(6, 1, 1)$ and a covariance matrix $1.8 \cdot I$. The original distribution of the datapoints is represented in the ``Vanilla'' facet of \cref{r-clarc-toy}.

We construct a 3-layer feed-forward neural network for binary classification with an input layer of 3 neurons, a hidden layer of 30 neurons, and an output layer of 2 neurons. The model is trained using the Adam optimizer with a learning rate of 0.01 for 5000 epochs. The decision boundary of the trained model is depicted as a light-blue surface in \cref{r-clarc-toy}.

We compute pattern-\gls{cav}s and train two linear SVM classifiers with L2 regularization and squared hinge loss for the classification of the two artifacts. Subsequently, we apply three distinct transformations to the original data: \gls{pclarc} (\cref{eq:mp-clarc}), class-conditional \gls{rclarc} (\cref{eq:r-clarc}, \cref{eq:class-cond}), and artifact-conditional \gls{rclarc} (\cref{eq:r-clarc}, \cref{eq:art-cond}), where artifact-conditional \gls{rclarc} utilizes the linear SVM classifier for artifact presence detection. Instead of applying the transformation to activations, we directly apply them to the input data. \cref{r-clarc-toy} illustrates these transformations.

In all ClArC transformations, only the data points are altered, while the decision boundary remains unchanged, as the model weights remain constant. We observe that \gls{pclarc} uniformly transforms all samples, resulting in a significant number of Class 2 data points (orange squares) crossing the decision boundary and leading to misclassification. In contrast, class-conditional \gls{rclarc} preserves model accuracy by leaving unchanged the data points classified by the model as Class 2. Artifact-conditional \gls{rclarc} exclusively transforms data points classified by the SVM classifiers as Artifact 1, Artifact 2, or both, further preserving the original data distribution while maintaining accuracy.

\begin{figure*}[ht]
  \includegraphics[width=\textwidth]{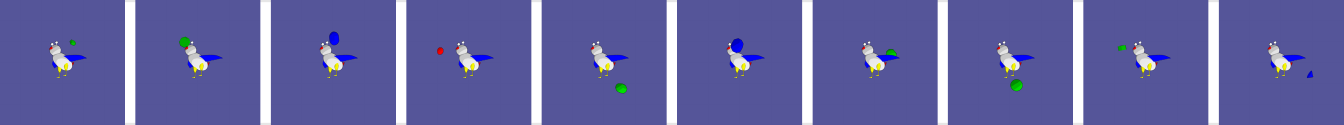}
  \centering
  \caption{10 artifacts included in Class 0 of FunnyBirds \emph{shortcut} dataset. The positions of the 10 background objects relative to the bird objects within the scene are fixed.}
\label{10-artifacts}
\end{figure*}

\section{Experimental Details} \label{exp_details_appx}

We outline the details of the two generated datasets for FunnyBirds, as well as the original and poisoned datasets for ISIC2019, in \cref{datasets}. \Cref{train} provides insights into the training process of Vanilla models. Information about the \gls{cav} calculation method is covered in \cref{cav_details}, while we evaluate \gls{cav}s in \cref{cav_eval}. Lastly, \cref{correction} outlines the details of the P- and \gls{rclarc}s model correction methods and their evaluation.

\subsection{Datasets} \label{datasets}

The FunnyBirds framework~\cite{hesse_funnybirds_2023} provides a framework for the creation of controlled datasets featuring 3D-rendered birds. Each bird class comprises 5 parts, with multiple options available for each part (\eg, 4 beaks, 3 eyes, etc.), which are assembled to form a bird sample. These samples are then placed within 3D scenes, where parameters such as camera position, zoom, lighting, and background objects are randomly selected for each sample.

We generated the \emph{backdoor} FunnyBirds dataset, which comprises 2 classes of birds. The defining parts for the two classes were randomly selected. As a backdoor artifact, we randomly selected the ``green box'' background object (see \cref{fig:generated}) and predefined its position relative to the bird's position within the 3D scene's coordinate system. For training and validation, we created a dataset consisting of 5000 samples of each of the two classes. 33\% of the labels of samples of class 0 were flipped to encourage learning the backdoor artifact. A randomly chosen 10\% subset was allocated for validation. Additionally, we constructed a test set comprising 100 correctly labeled birds from each class.

The \emph{shortcut} FunnyBirds dataset comprises 10 different classes. We incentivized the utilization of the shortcut artifact by designing classes 0 to 3 to only vary in the beak part; otherwise, the parts of other classes were randomly selected. We generated 10 different background object artifacts with predetermined positions relative to the bird (as depicted in \cref{10-artifacts}), which were inserted into samples of class 0. Specifically, 50\% of class 0 bird samples included a randomly selected number from 1 to 10 of these artifacts. We ensured an equal distribution of the total number of background artifacts between shortcut and non-shortcut samples by setting the minimum number of background objects for each sample to 10. This way, we ensured that the number of background objects was not used as a spurious feature. We generated a dataset with 1000 birds of each class, with 10\% of this set allocated for validation. Additionally, we constructed a test set comprising 100 birds from each class.

For both FunnyBirds datasets, we generated binary masks that precisely localize the artifact object using the functionality of the FunnyBirds framework. These binary masks are employed to assess artifact relevance in \cref{performance}.

\begin{figure}[ht]
  \includegraphics[width=\columnwidth]{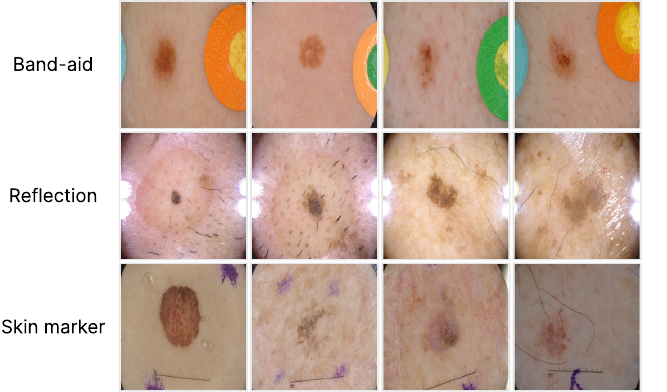}
  \centering
  \caption{Examples of ISIC artifacts band-aid (``NV''), skin marker (``NV'', ``BKL''), and reflection (``BKL'') artifacts.}
\label{isic_exmples}
\end{figure}

The ISIC2019~\cite{8363547,combalia2019bcn20000, mitsuhara2019embedding} dataset consists of 25,331 samples of classes ``MEL'', ``NV'', ``BCC'', ``AK'', ``BKL'', ``DF'', ``VASC'', and ``SCC''. We used the Reveal2Revise (R2R) framework~\cite{pahde_reveal_2023} to identify three artifacts naturally occurring in the dataset, strongly correlating with class labels: band-aid (``NV''), skin marker (``NV'', ``BKL''), and reflection (``BKL'') artifacts (examples are provided in \cref{isic_exmples}). Following the R2R approach, we identified artifact samples and computed artifact localization binary masks. Firstly, we obtained artifact localization heatmaps by generating Layer-wise Relevance Propagation (LRP)~\cite{10.1371/journal.pone.0130140} heatmaps for SVM-\gls{cav}s~\cite{kim_interpretability_2018} in the features.7 layer of VGG16 trained on ISIC2019 data (training details of the model were outlined in \cref{train}). The LRP attribution heatmaps are computed using the $\epsilon z^{+} \flat$-composite~\cite{kohlbrenner2020towards} with the \texttt{zennit} library~\cite{anders2021zennit}. Secondly, we manually sorted the heatmaps to exclude those that appeared to have high attributions in regions unrelated to the artifact concept. Thirdly, pixels corresponding to positive attributions (larger than $\epsilon=0.3$) received a value of 1 in the binary masks, while others were assigned a value of 0. These artifact binary localization masks were subsequently utilized in evaluating artifact relevance (as described in \cref{performance}). Additionally, we utilized the \gls{cav} heatmaps to isolate the artifacts in the corresponding samples and overlay them onto clean test samples for the ISIC2019 ``poisoned'' setting and for the ``generated'' \gls{cav} datasets \cref{cav_calculation}. Specifically, the artifact sample image was multiplied with its \gls{cav} heatmap element-wise and then added to the clean sample with the pixel values multiplied by (1 - attribution) element-wise.

\begin{table}[htbp]
\centering
\small
\begin{tabularx}{\columnwidth}{p{0.08\textwidth}p{0.12\textwidth}p{0.07\textwidth}YY}
\hline
\multicolumn{1}{c}{Dataset} & \multicolumn{1}{c}{Model} & \multicolumn{1}{c}{Optimizer} & \multicolumn{1}{c}{LR} & \multicolumn{1}{c}{Epochs} \\
\hline
\multirow{3}{*}{ISIC2019} & VGG16 & SGD & 0.05 & 150 \\
 & ResNet18 & SGD & 0.05 & 150 \\
 & EfficientNet-B0 & Adam & 0.01 & 150 \\
 \hline
\multirow{3}{*}{\specialcell{FunnyBirds\\\emph{backdoor}}} & VGG16 & SGD & 0.001 & 100 \\
 & ResNet18 & Adam & 0.001 & 100 \\
 & EfficientNet-B0 & Adam & 0.001 & 100 \\
\hline
\multirow{3}{*}{\specialcell{FunnyBirds\\\emph{shortcut}}} & VGG16 & SGD & 0.001 & 100 \\
 & ResNet18 & Adam & 0.001 & 100 \\
 & EfficientNet-B0 & Adam & 0.001 & 100 \\
\hline
\end{tabularx}
\caption{Model training details including optimizer, initial learning Rate (LR), number of epochs.}
\label{tab:dataset-models}
\end{table}

\begin{table*}[t]
    \centering
    \small
    \begin{tabularx}{\textwidth}{p{0.15\textwidth}p{0.10\textwidth}YYYYYYYY}
\toprule 
        \multirow{3}{*}{Model} & \multirow{3}{*}{\gls{cav} Dataset} & \multicolumn{2}{c}{FunnyBirds} & \multicolumn{6}{c}{ISIC2019} \\
             \cmidrule(lr){3-4} \cmidrule(lr){5-10}
        &  & \multicolumn{2}{c}{``green box''} & \multicolumn{2}{c}{``reflection''} & \multicolumn{2}{c}{``band-aid''}& \multicolumn{2}{c}{``skin marker''}\\
        &  & \multicolumn{1}{c}{Pattern} & \multicolumn{1}{c}{Filter}& \multicolumn{1}{c}{Pattern} & \multicolumn{1}{c}{Filter}& \multicolumn{1}{c}{Pattern} & \multicolumn{1}{c}{Filter}& \multicolumn{1}{c}{Pattern} & \multicolumn{1}{c}{Filter}\\
        
        \midrule
\multirow[c]{2}{*}{ResNet18} & Generated & $0.824$ & $0.101$ & $0.617$ & $0.343$ & $0.406$ & $0.215$ & $0.406$ & $0.166$ \\
 & Data Subset & $0.563$ & $0.042$ & $0.469$ & $0.328$ & $0.166$ & $0.156$ & $0.193$ & $0.089$ \\
\cmidrule{1-10}
\multirow[c]{2}{*}{VGG16} & Generated & $0.779$ & $0.132$ & $0.919$ & $0.180$ & $0.588$ & $0.132$ & $0.608$ & $0.128$ \\
 & Data Subset & $0.345$ & $0.119$ & $0.885$ & $0.430$ & $0.482$ & $0.235$ & $0.443$ & $0.121$ \\
\cmidrule{1-10}
\multirow[c]{2}{*}{EfficientNet-B0} & Generated & $0.859$ & $0.040$ & $0.655$ & $0.093$ & $0.454$ & $0.054$ & $0.440$ & $0.088$ \\
 & Data Subset & $0.854$ & $0.002$ & $0.332$ & $0.424$ & $-0.077$ & $0.080$ & $-0.012$ & $0.056$ \\
 \bottomrule
    \end{tabularx}
        \caption{Evaluating the alignment of artifact \glspl{cav} in terms of cosine similarity with the actual change in activations when the concept is added in a controlled fashion across various models, the FunnyBirds backdoor dataset, and the ISIC2019 dataset. We compare \glspl{cav} computed on original data subsets, and \gls{cav}s computed on pairs of clean and (generated) poisoned samples.}
    \label{tab:cav_alignment_BIG}
\end{table*}

\begin{table*}[t]
    \centering
    \small
    \setlength{\tabcolsep}{5pt}
\begin{tabularx}{\textwidth}{p{0.15\textwidth}cccccccccccccc}
\toprule
        \multirow{2}{*}{Model} & \multicolumn{1}{c}{\specialcell{FunnyBirds\\ \emph{backdoor}}} &  \multicolumn{10}{c}{\specialcell{FunnyBirds\\ \emph{shortcut}}} & \multicolumn{3}{c}{ISIC2019}\\
            \cmidrule(lr){2-2} \cmidrule(lr){3-12} \cmidrule(lr){13-15}
          & \multicolumn{1}{c}{``GB''} & \multicolumn{1}{c}{0} & \multicolumn{1}{c}{1} & \multicolumn{1}{c}{2} & \multicolumn{1}{c}{3} & \multicolumn{1}{c}{4} & \multicolumn{1}{c}{5} & \multicolumn{1}{c}{6} & \multicolumn{1}{c}{7} & \multicolumn{1}{c}{8} & \multicolumn{1}{c}{9} & \multicolumn{1}{c}{``R''} & \multicolumn{1}{c}{``BA''}& \multicolumn{1}{c}{``SM''}\\
\midrule
ResNet18 & $94.1$ & $96.3$ & $96.2$ & $94.0$ & $95.6$ & $94.9$ & $97.5$ & $96.3$ & $94.9$ & $95.6$ & $96.3$ & $98.4$ & $95.9$ & $100.0$ \\
VGG16 & $90.4$ & $94.4$ & $93.1$ & $94.0$ & $93.1$ & $88.6$ & $95.7$ & $93.9$ & $93.7$ & $93.1$ & $93.2$ & $93.4$ & $100.0$ & $96.7$ \\
EfficientNet-B0 & $90.1$ & $93.8$ & $93.1$ & $93.4$ & $93.8$ & $91.8$ & $94.4$ & $92.6$ & $94.3$ & $94.4$ & $93.8$ & $91.8$ & $95.9$ & $100.0$ \\
\bottomrule
    \end{tabularx}
        \caption{Hold-out set accuracies of linear SVM classifiers across diverse datasets, artifacts, and models. These classifiers serve as artifact-condition-generating functions in artifact-conditional and combined \gls{rclarc}.  ``GB'' denotes ``green box'', ``R'' represents ``reflection'', ``BA'' indicates ``band-aid'', and ``SM'' signifies ``skin marker''.}
    \label{tab:SVM_acc}
\end{table*}

\subsection{Model Training} \label{train}

\cref{tab:dataset-models} provides training details for all models and datasets, including optimizer, initial learning rate (LR), and number of epochs. The ISIC2019 models were pre-trained on ImageNet~\cite{russakovsky_imagenet_2015} using weights obtained from the \texttt{Torchvision} library. The learning rate (LR) for the ISIC2019 model was divided by 10 after epochs 50 and 80 during training. Both FunnyBirds models were trained from scratch, employing early stopping based on validation set loss with a patience of 3 epochs.

\subsection{\gls{cav} Calculation} \label{cav_details}

For both FunnyBirds datasets, we create respectively additional 1000 negative samples of class 0 birds, as in both cases this class is associated with artifacts. In the \emph{backdoor} FunnyBirds dataset, we then generate 1000 images with the ``green box'' artifact, while for the  ``shortcut'' FunnyBirds dataset, we produce a set of 1000 positive examples for each of the 10 artifacts.

To create negative example sets for ISIC2019 for the generated \gls{cav}s, we begin by sampling 1000 non-artifact images from the classes associated with each artifact. Subsequently, we overlay the cropped-out artifacts onto these images, following the process outlined in \cref{datasets}, resulting in a set of 1000 positive examples for each artifact. For dataset subset \gls{cav}s, we use all available artifact samples as positive examples and sample negative examples from the ISIC2019 non-artifact samples. We ensure that the ratio of positive to negative examples does not exceed 5.

For pattern-based \gls{cav} calculation we directly adopt the approach from~\cite{pahde_navigating_2024}, while for filter-based \gls{cav}s we employ linear SVMs trained with L2 regularization and squared hinge loss with class weights inversely proportional to class frequencies.

\subsection{\gls{cav} Evaluation} \label{cav_eval}

The alignment scores for various \gls{cav}s methods were computed following the approach outlined in~\cite{pahde_navigating_2024}. We present the \gls{cav} evaluation results for different model architectures, for the FunnyBirds backdoor ``green box'' artifact, and all examined ISIC2019 artifacts in \cref{tab:cav_alignment_BIG}.

\subsection{Model Correction and Evaluation} \label{correction}

We assess artifact relevance using heatmaps computed with LRP using the $\epsilon z^{+} \flat$-composite~\cite{kohlbrenner2020towards} with the \texttt{zennit} library~\cite{anders2021zennit}. The procedure for generating binary localization masks is detailed in \cref{datasets}. Artifact relevance is quantified as the sum of absolute attribution values within the mask divided by the sum of all absolute attribution values. To visualize the LRP heatmaps in \cref{fig:heatmaps}, we normalize them by dividing each heatmap by its maximum absolute value. 

Our evaluation encompasses \gls{pclarc}, as well as \gls{rclarc} with class-conditional and artifact-conditional condition-generating functions, along with their combination. Model correction is performed for all models post the last convolutional layer, utilizing pattern-based ``generated'' \gls{cav}s. For artifact-conditional and combined \gls{rclarc}s, our artifact-conditioning function classifies the samples in the latent space of the last convolutional layer as well. For this, we employ linear SVMs trained with L2 regularization and squared hinge loss, with class weights inversely proportional to class frequencies. The training data consists of all available artifact samples as positive examples and a subset of negative examples from the dataset, ensuring the positive-to-negative example ratio does not exceed 5. 20\% of the training set serves as a holdout set to assess classifier accuracy. The accuracies of the resulting SVM classifiers are presented in \cref{tab:SVM_acc}.

\section{Further Experiments} \label{more-eval}

\begin{figure*}[htbp]
\begin{subfigure}{0.5\textwidth}
  \includegraphics[width=\linewidth]{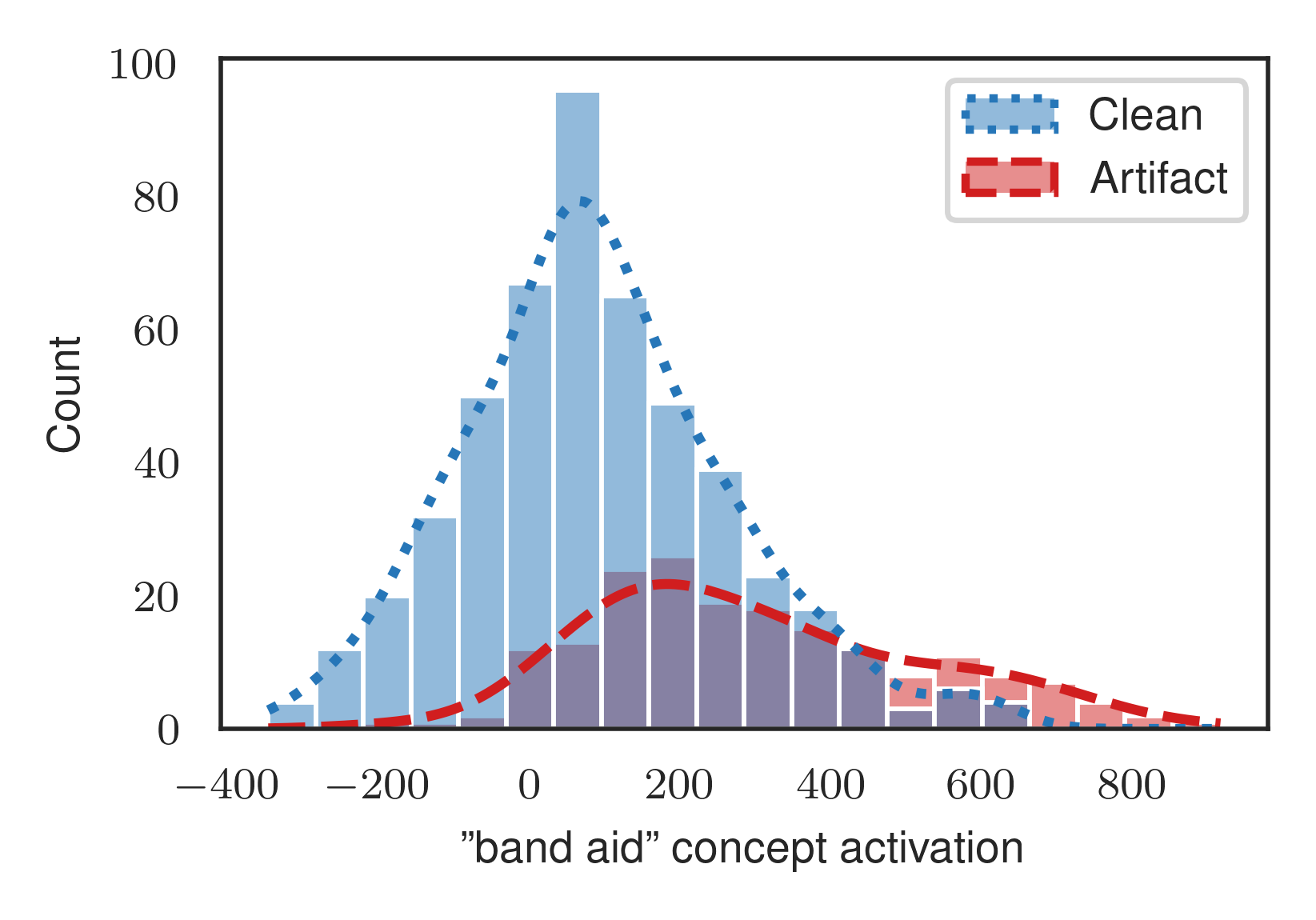}
  \caption{``Band-Aid''}
  \label{fig:todo1}
\end{subfigure}
\begin{subfigure}{0.5\textwidth}
  \includegraphics[width=\linewidth]{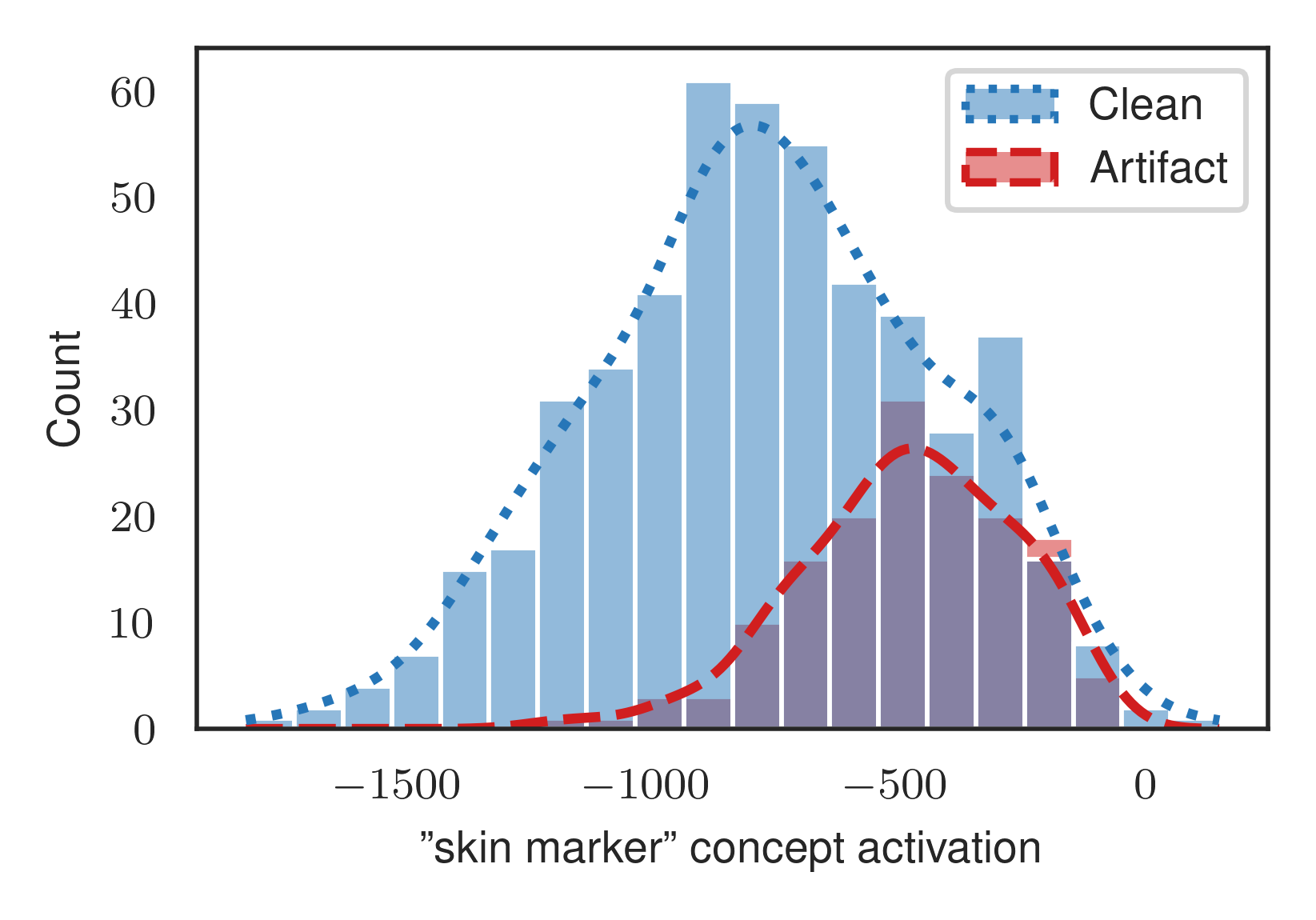}
  \caption{``Skin Marker''}
  \label{fig:todo2}
\end{subfigure}
\caption{Histogram illustrating the activations of artifact \gls{cav}s for the corresponding artifact samples alongside 500 randomly selected clean samples for ISIC2019 dataset and ResNet18 model.}
\label{fig:combined}
\end{figure*}

\begin{figure*}[ht]
  \includegraphics[width=\textwidth]{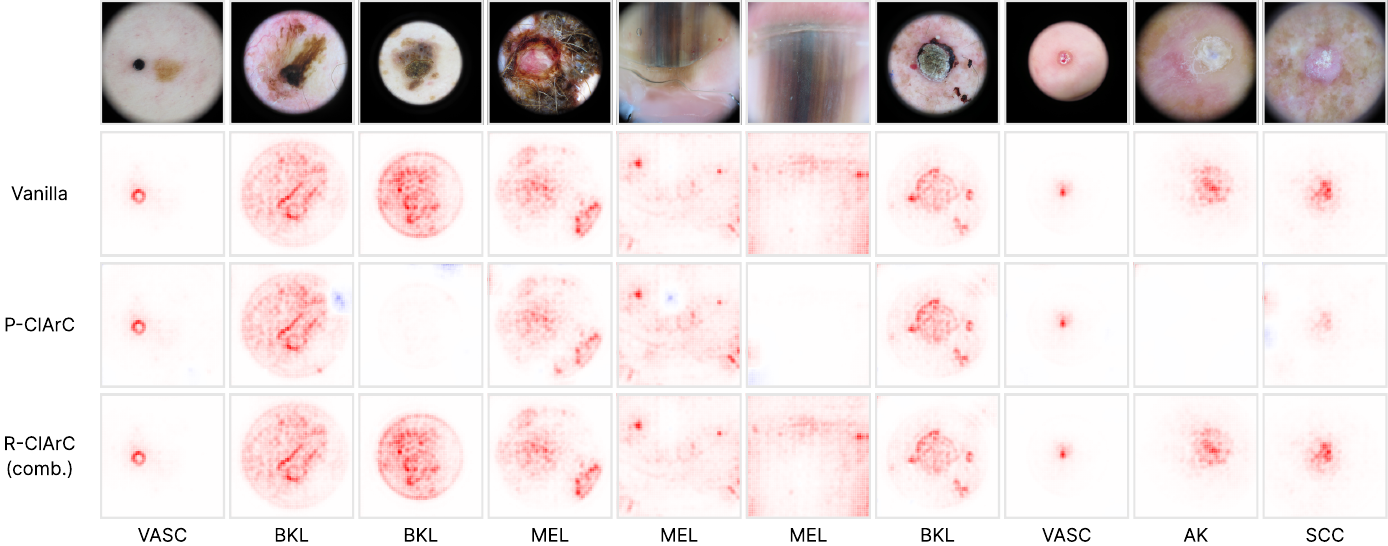}
  \centering
  \caption{LRP heatmaps depicting samples with pronounced activation of the ``reflection'' concept for the Vanilla model and models corrected using \gls{pclarc} and \gls{rclarc} combining class- and artifact-conditional approaches.}
\label{todo3}
\end{figure*}

\begin{figure*}[ht]
  \includegraphics[width=\textwidth]{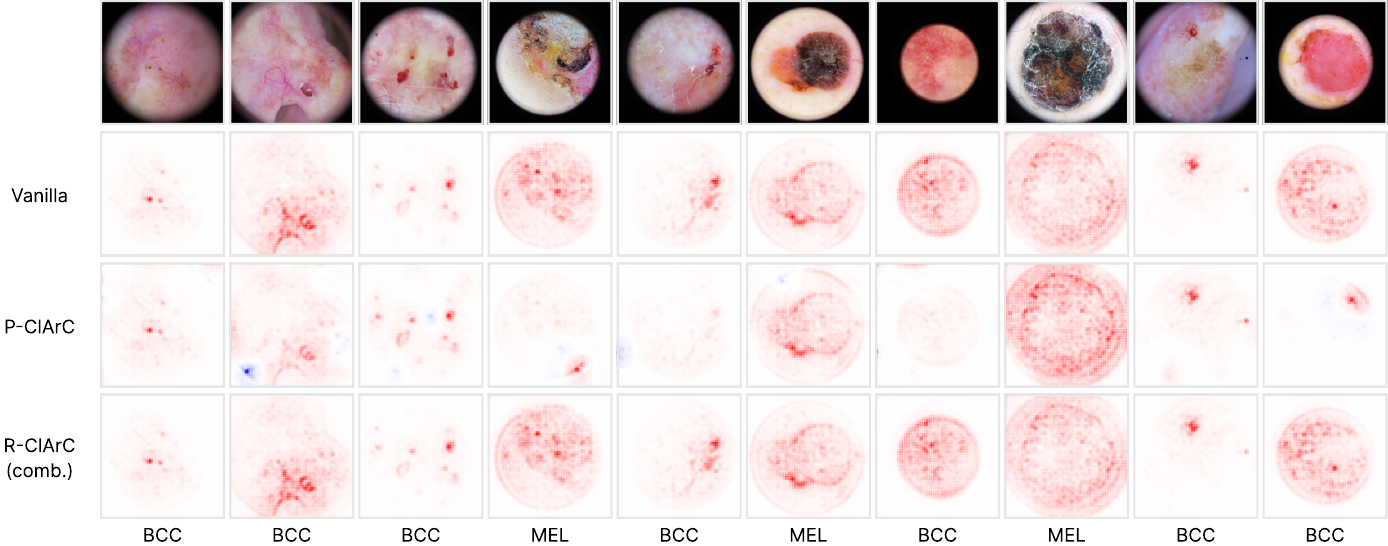}
  \centering
  \caption{LRP heatmaps depicting samples with pronounced activation of the ``band-aid'' concept for the Vanilla model and models corrected using \gls{pclarc} and \gls{rclarc} combining class- and artifact-conditional approaches.}
\label{todo4}
\end{figure*}

\begin{figure*}[ht]
  \includegraphics[width=\textwidth]{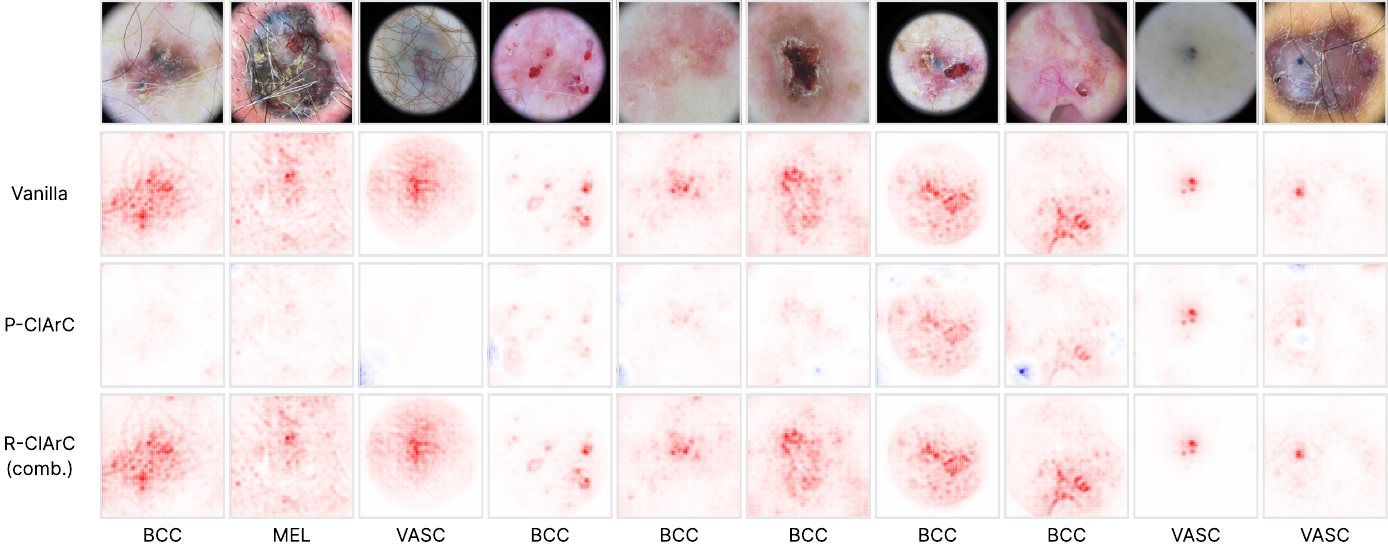}
  \centering
  \caption{LRP heatmaps depicting samples with pronounced activation of the ``skin marker'' concept for the Vanilla model and models corrected using \gls{pclarc} and \gls{rclarc} combining class- and artifact-conditional approaches.}
\label{todo5}
\end{figure*}

In the following section, we present supporting experiments aimed at testing the orthogonality of concepts (\cref{tab:10_art_en}, \cref{fig:combined}). Additionally, we provide further heatmaps to facilitate qualitative evaluation of the \gls{rclarc} method compared to \gls{pclarc} (\cref{todo3}, \cref{todo4}, \cref{todo5}).

\begin{table*}[t]
    \centering
    \small
\begin{tabularx}{\textwidth}{YYYYYYYYYYY}
\toprule
Artifacts & Class 0 & Class 1 & Class 2 & Class 3 & Class 4 & Class 5 & Class 6 & Class 7 & Class 8 & Class 9 \\
\midrule
0 & $-0.33$ & $0.12$ & $-0.65$ & $0.62$ & $0.40$ & $0.45$ & $0.34$ & $-0.75$ & $0.09$ & $0.19$ \\
1 & $0.59$ & $0.07$ & $0.79$ & $-0.17$ & $-0.78$ & $-0.59$ & $-0.77$ & $0.47$ & $0.48$ & $0.01$ \\
2 & $0.51$ & $-0.15$ & $0.80$ & $-0.53$ & $-0.73$ & $-0.44$ & $-0.64$ & $0.70$ & $0.40$ & $-0.23$ \\
3 & $-0.45$ & $-0.08$ & $-0.78$ & $0.53$ & $0.61$ & $0.62$ & $0.57$ & $-0.60$ & $-0.23$ & $0.01$ \\
4 & $0.13$ & $0.04$ & $0.31$ & $-0.31$ & $-0.46$ & $-0.12$ & $-0.32$ & $0.07$ & $0.62$ & $0.00$ \\
5 & $-0.23$ & $0.37$ & $-0.45$ & $0.54$ & $0.29$ & $0.12$ & $0.20$ & $-0.74$ & $0.09$ & $0.43$ \\
6 & $0.54$ & $0.13$ & $0.77$ & $-0.26$ & $-0.73$ & $-0.57$ & $-0.74$ & $0.49$ & $0.37$ & $0.07$ \\
7 & $-0.27$ & $0.13$ & $-0.03$ & $-0.28$ & $-0.11$ & $0.13$ & $0.05$ & $-0.13$ & $0.33$ & $0.12$ \\
8 & $0.70$ & $-0.27$ & $0.90$ & $-0.44$ & $-0.80$ & $-0.51$ & $-0.76$ & $0.84$ & $0.35$ & $-0.35$ \\
9 & $-0.36$ & $0.43$ & $-0.54$ & $0.49$ & $0.48$ & $0.10$ & $0.33$ & $-0.73$ & $-0.13$ & $0.49$ \\
\bottomrule
    \end{tabularx}
        \caption{Cosine similarity between artifact \gls{cav}s and the mean feature direction of each class for the FunnyBirds shortcut dataset and EfficientNet-B0. The strong relationship between artifact and class direction explains the strong negative impact of ClArC transformations on model performance. Suppressing the artifact direction results in pushing samples across the decision boundary.}
    \label{tab:10_art_en}
\end{table*}

\clearpage

\end{document}